\documentclass[letterpaper,11pt]{article}

\usepackage[margin=1in]{geometry}
\usepackage[singlespacing]{setspace}

\usepackage[utf8]{inputenc} 
\usepackage[T1]{fontenc}    
\usepackage{natbib}
\usepackage{enumitem}
\usepackage{xspace}
\usepackage{hyperref}       
\hypersetup{
    colorlinks,
    breaklinks,
    linkcolor = blue,
    citecolor = blue,
    urlcolor  = blue,
}
\usepackage{url}            
\usepackage{amsfonts}       
\usepackage{nicefrac}       
\usepackage{microtype}      
\usepackage{amsmath,bm}
\usepackage{amsthm}
\usepackage[algo2e, ruled, vlined]{algorithm2e}
\usepackage{algorithmicx}
\usepackage{algorithm}
\usepackage[noend]{algcompatible}
\usepackage{bbm}      
\usepackage{mathtools}  \usepackage{matlab-prettifier}
\usepackage{amsopn}
\usepackage{amssymb}
\usepackage{graphicx}
\usepackage{xcolor}
\usepackage{footnote}
\usepackage{makecell}
\usepackage{multirow}
\makesavenoteenv{tabular}
\makesavenoteenv{table}
\usepackage[table]{xcolor}
\usepackage{booktabs}
\usepackage{thmtools}
\usepackage{thm-restate}
\makeatletter
\@ifundefined{newcounteralias}{}{%
  \renewcommand\thmt@autorefsetup{%
    \@xa\def\csname\thmt@envname autorefname
    \@xa\endcsname\@xa{\thmt@thmname}%
  }%
}
\makeatother

\usepackage{xcolor}
\usepackage{soul}
\usepackage{changepage}

\definecolor{Green}{rgb}{0.13, 0.65, 0.3}
\definecolor{Amber}{rgb}{0.3, 0.5, 1.0}
\usepackage[]{color-edits}
\addauthor{HL}{Green}
\addauthor{TJ}{Amber}
\usepackage{comment}

\sethlcolor{gray}

\newcommand{\KLComp}{\textsc{KL-Comp}}

\newcommand{\Ind}[1]{ \field{I}{\left\{{#1}\right\}} }

\newcommand{\Ber}{\texttt{Ber}}

\newcommand{\bi}{\begin{itemize}}
\newcommand{\ei}{\end{itemize}}

\theoremstyle{plain} 
\newtheorem{theorem}{Theorem}[section]
\newtheorem{lemma}[theorem]{Lemma}

\newtheorem{proposition}[theorem]{Proposition}

\usepackage{amssymb}
\usepackage{pifont}%

\allowdisplaybreaks

\DeclareMathOperator*{\argmin}{\arg\!\min}

\usepackage{amsmath}
\usepackage{amssymb}
\usepackage{graphicx}
\usepackage{mathtools}
\usepackage{footnote}
\usepackage{float}
\usepackage{xspace}
\usepackage{multirow}
\usepackage{nicefrac}
\usepackage{wrapfig}
\usepackage{framed}
\usepackage{url}
\usepackage{tikz}
\usetikzlibrary{arrows,chains,matrix,positioning,scopes}

\newcommand{\calL}{{\mathcal{L}}}

\newcommand{\calX}{{\mathcal{X}}}

\newcommand{\calF}{{\mathcal{F}}}

\newcommand{\field}[1]{\mathbb{#1}}

\newcommand{\E}{\field{E}}

\newcommand{\KL}{{\text{\rm KL}}}

\newcommand{\Reg}{{\text{\rm Reg}}}

\newcommand{\order}{\ensuremath{\mathcal{O}}}
\newcommand{\otil}{\ensuremath{\widetilde{\mathcal{O}}}}

\newcommand{\rbr}[1]{\left(#1\right)}

\newcommand{\sbr}[1]{\left[#1\right]}

\newcommand{\cbr}[1]{\left\{#1\right\}}

\DeclareFontFamily{OMX}{MnSymbolE}{}
\DeclareFontShape{OMX}{MnSymbolE}{m}{n}{
    <-6>  MnSymbolE5
   <6-7>  MnSymbolE6
   <7-8>  MnSymbolE7
   <8-9>  MnSymbolE8
   <9-10> MnSymbolE9
  <10-12> MnSymbolE10
  <12->   MnSymbolE12}{}
\DeclareSymbolFont{mnlargesymbols}{OMX}{MnSymbolE}{m}{n}
\SetSymbolFont{mnlargesymbols}{bold}{OMX}{MnSymbolE}{b}{n}
\DeclareMathDelimiter{\llangle}{\mathopen}{mnlargesymbols}{'164}{mnlargesymbols}{'164}
\DeclareMathDelimiter{\rrangle}{\mathclose}{mnlargesymbols}{'171}{mnlargesymbols}{'171}

\usepackage{prettyref}
\newcommand{\pref}[1]{\prettyref{#1}}

\newcommand{\savehyperref}[2]{\texorpdfstring{\hyperref[#1]{#2}}{#2}}
\newrefformat{eq}{\savehyperref{#1}{Eq.~\eqref{#1}}}
\newrefformat{eqn}{\savehyperref{#1}{Equation~\eqref{#1}}}
\newrefformat{lem}{\savehyperref{#1}{Lemma~\ref*{#1}}}
\newrefformat{def}{\savehyperref{#1}{Definition~\ref*{#1}}}
\newrefformat{line}{\savehyperref{#1}{line~\ref*{#1}}}
\newrefformat{thm}{\savehyperref{#1}{Theorem~\ref*{#1}}}
\newrefformat{cond}{\savehyperref{#1}{Condition~\ref*{#1}}}
\newrefformat{corr}{\savehyperref{#1}{Corollary~\ref*{#1}}}
\newrefformat{cor}{\savehyperref{#1}{Corollary~\ref*{#1}}}
\newrefformat{col}{\savehyperref{#1}{Corollary~\ref*{#1}}}
\newrefformat{sec}{\savehyperref{#1}{Section~\ref*{#1}}}
\newrefformat{app}{\savehyperref{#1}{Appendix~\ref*{#1}}}
\newrefformat{assum}{\savehyperref{#1}{Assumption~\ref*{#1}}}
\newrefformat{ex}{\savehyperref{#1}{Example~\ref*{#1}}}
\newrefformat{fig}{\savehyperref{#1}{Figure~\ref*{#1}}}
\newrefformat{tab}{\savehyperref{#1}{Table~\ref*{#1}}}
\newrefformat{alg}{\savehyperref{#1}{Algorithm~\ref*{#1}}}
\newrefformat{rem}{\savehyperref{#1}{Remark~\ref*{#1}}}
\newrefformat{conj}{\savehyperref{#1}{Conjecture~\ref*{#1}}}
\newrefformat{prop}{\savehyperref{#1}{Proposition~\ref*{#1}}}
\newrefformat{proto}{\savehyperref{#1}{Protocol~\ref*{#1}}}
\newrefformat{fn}{\savehyperref{#1}{Footnote~\ref*{#1}}}
\newrefformat{prob}{\savehyperref{#1}{Problem~\ref*{#1}}}
\newrefformat{claim}{\savehyperref{#1}{Claim~\ref*{#1}}}
\newrefformat{que}{\savehyperref{#1}{Question~\ref*{#1}}}
\newrefformat{obs}{\savehyperref{#1}{Observation~\ref*{#1}}}

\newcommand{\calLbd}{\mathcal{L}_{\mathrm{bd}}}

\newcommand{\calLmix}{\mathcal{L}_{\mathrm{mix}}}

\newcommand{\dmeas}{\,\mathrm{d}}

\newcommand{\Unif}{\operatorname{Unif}}

\newcommand{\LipLosses}[1]{\mathcal{L}_{\mathrm{Lip}}\rbr{#1}}
\newcommand{\MixLosses}[1]{\mathcal{L}_{\mathrm{mix}}\rbr{#1}}

\newcommand{\Leb}{\operatorname{Leb}}
\newcommand{\ClipOp}{\operatorname{clip}}

\usepackage{hyperref}
\hypersetup{
  colorlinks=true,
  breaklinks=true,
  linkcolor=blue,
  citecolor=blue,
  urlcolor=blue
}

\title{Omniprediction for All Mixable Proper Losses and Beyond\thanks{Authors are listed in alphabetical order.}}
\author{%
Junyan Liu\thanks{University of Washington. Email: \texttt{junyanl1@cs.washington.edu}.}
\and
Haipeng Luo\thanks{University of Southern California. Email: \texttt{haipengl@usc.edu}.}
\and
Lillian J. Ratliff\thanks{University of Washington. Email: \texttt{ratliffl@uw.edu}.
}
}

\date{}

\begin{document}
\maketitle
\thispagestyle{empty}
\vspace{-2em}

\begin{abstract}
  Omniprediction concerns the task of making predictions that are simultaneously accurate with respect to different losses.
  Earlier results by~\cite{chernov2009prediction} show that in an online learning setting,
  $\order(\log(|\calF||\calL|))$ regret is achievable simultaneously against every predictor in a finite predictor class $\calF$ and for every loss in a finite class $\calL$ of mixable proper losses,
  while recent advances~\citep{okoroafor2025near, farina2026blackbox} show that $\otil(\sqrt{T\log(|\calF||\calL|)})$ regret is achievable simultaneously for every loss in a finite class $\calL$ of bounded losses.

  In this work, we ask whether sublinear regret remains possible when $\calL$ is an \textit{infinite} class of proper losses.
  Our first result gives a negative answer in general: even when $\calF$ contains only two linear predictors, it is impossible to achieve sublinear regret simultaneously for all bounded proper losses.
  We then show that $\otil(T^{1/3}(\log|\calF|)^{2/3})$ regret is achievable simultaneously for \textit{all} mixable proper losses and that this bound is optimal up to logarithmic factors.
  This optimal rate stands in stark contrast to the constant (in $T$) regret achievable for a finite class of mixable proper losses.
  Finally, we extend our algorithm to obtain a best-of-all-worlds result: in addition to the new regret guarantee for all mixable proper losses, the resulting algorithm simultaneously achieves the existing guarantees for finite loss classes.
  With appropriate choices of these finite classes, our algorithm simultaneously achieves constant regret for squared loss and log loss, $\otil(T^{1/3}(\log|\calF|)^{2/3})$ regret for all mixable proper losses,
  and $\otil(\sqrt{T\log|\calF|})$ regret for $0$-$1$ loss, all  Lipschitz losses, all GLM losses,
  and other structured families.
\end{abstract}


\clearpage
\tableofcontents
\thispagestyle{empty}

\clearpage
\pagenumbering{arabic}

\section{Introduction}
\label{sec:intro}
A standard machine learning approach to making predictions is to choose a loss function and train a predictor to minimize it.
In many applications, however, the same predictions inform multiple downstream decisions, each with a different loss function, and these losses may not be known at training time.
\emph{Omniprediction} addresses this mismatch by seeking a single predictor whose outputs can be post-processed to compete with a benchmark class $\calF$ simultaneously for every loss in a class $\calL$.
The post-processing treats the predictor's output as the true outcome probability and chooses a Bayes-optimal action for the desired loss.
Thus, rather than training a separate model for each objective, omniprediction separates the task of forecasting from the choice of how to act on a forecast.
This paradigm was introduced in the batch setting by~\cite{gopalan2022omnipredictors} and subsequently extended to the online adversarial setting by~\cite{garg2024oracle}.

In the online setting, the learner must produce forecasts sequentially, without knowing future outcomes, and the goal is to achieve sublinear cumulative regret against the best predictor in $\calF$ for every loss of interest.
More specifically, we consider the following contextual online setting for forecasting binary outcomes:
for each round $t=1, \ldots, T$, the nature reveals a context $x_t$ from an arbitrary context space $\calX$, the learner then makes a prediction $P_t \in [0,1]$, and finally the nature reveals the true outcome $y_t\in\cbr{0,1}$.
Here, $P_t$ represents the learner's forecast on the probability of $y_t$ being $1$ and is generally random; we let $D_t$ be the distribution over $[0,1]$ from which $P_t$ is drawn.
For a benchmark class $\calF \subseteq [0,1]^\calX$ and a loss function
$\ell:[0,1]\times\cbr{0,1}\to\mathbb R   
\cup\{+\infty\}$, the regret of the learner is defined as:
\[
  \Reg_{\ell,\calF}(T)
  =
  \sum_{t=1}^T
  \E_{P_t\sim D_t}\sbr{\ell\rbr{\pi_\ell(P_t),y_t}}
  -
  \inf_{f\in\calF}
  \sum_{t=1}^T\ell\rbr{f(x_t),y_t},
\]
where $\pi_\ell(p) = \argmin_{q\in[0,1]} \E_{Y\sim\Ber(p)}\sbr{\ell(q,Y)}$ (with $\Ber(p)$ being the Bernoulli distribution with mean $p$ and tie broken in an arbitrary, fixed way) is the post-processing function that returns one of the best actions based on the learner's forecast on the outcome and the loss function.
Therefore, regret $\Reg_{\ell,\calF}(T)$ can be interpreted as the difference between the total loss of a downstream agent who consumes the learner's prediction and acts optimally and that of the best fixed benchmark from the class $\calF$.
Since different agents might care about different loss functions, the goal of omniprediction is to make sure $\Reg_{\ell,\calF}(T)$ is small simultaneously for all $\ell \in \calL$ where $\calL$ is a certain broad class of loss functions.

Although the terminology of omniprediction is recent, earlier work in fact already established strong simultaneous regret guarantees for certain loss families.
In particular, \cite{chernov2009prediction} considered a finite class $\calL$ of \textit{proper} and \textit{mixable} losses:
\begin{itemize}
  \item (\textbf{Proper Losses}) A loss $\ell$ is proper if
    $
    \E_{Y\sim\Ber(p)}\sbr{\ell(p,Y)}
    \le
    \E_{Y\sim\Ber(p)}\sbr{\ell(q,Y)},
    \forall p,q\in[0,1],
    $
    that is, predicting $p$ is optimal when the outcome is indeed drawn from $\Ber(p)$.
    In this case we can simply let $\pi_\ell(p)=p$.

  \item (\textbf{Mixable Losses}) A loss $\ell$ is $\beta$-mixable if, for every distribution $D$ over $[0,1]$, there exists $p \in [0,1]$ such that
    $
    \exp\rbr{-\beta\ell(p,y)}
    \ge
    \E_{P\sim D}\exp\rbr{-\beta\ell(P,y)},
    \forall y\in\cbr{0,1}.
    $
    Denote by $\beta_\ell^\star$ the largest constant $\beta$ such that $\ell$ is $\beta$-mixable.
\end{itemize}

It is well-known that standard losses such as squared loss and log loss belong to this class; in particular, squared loss $\ell(p,y) = (p-y)^2$ is $2$-mixable while log loss $\ell(p,y) = -y\log p - (1-y)\log(1-p)$ is $1$-mixable~\citep{cesa2006prediction}.
Moreover, every $L$-Lipschitz proper loss is $2/L$-mixable (see \pref{lem:automatic-curvature}).
\cite{chernov2009prediction} shows that the defensive forecasting algorithm~\citep{chernov2010supermartingales} achieves $\Reg_{\ell,\calF}(T) = \order\rbr{\frac{\log(|\calF||\calL|)}{\beta_\ell^\star}}$ simultaneously for all $\ell$ from a finite class $\calL$ of proper and mixable loss functions.

On the other hand, recent advances by~\cite{okoroafor2025near, farina2026blackbox} achieve omniprediction guarantees beyond proper mixable losses:
their algorithms achieve
$
\Reg_{\ell,\calF}(T) = \otil\rbr{\sqrt{T\log(|\calF||\calL|)}}
$
simultaneously for all $\ell$ from a finite class $\calL$ of arbitrary, bounded loss functions.\footnote{Throughout this paper, $\otil(\cdot)$
hides only polylogarithmic factors in $T$.}
Importantly, these guarantees can also accommodate certain \emph{infinite} loss classes through finite approximate basis representations:
for example, with a suitable choice of basis and approximation precision, the guarantee above implies $\otil(L\sqrt{T\log|\calF|})$ regret for all $L$-Lipschitz losses
; see~\cite{okoroafor2025near} for details as well as \pref{tab:summary} for a summary.

However, not every loss family admits a finite approximation basis.
A particularly important example is the class of all bounded proper losses.
Indeed, as pointed out by~\cite{okoroafor2025near},
proper losses can have discontinuities at arbitrary thresholds, and the threshold basis used to represent the full class cannot simply be replaced by a fixed finite discretization (unless one restricts $\calF$ to be a class with finitely many possible outputs, which is an unnatural assumption).
This leads to the first main question our work aims to address:
\begin{quote}
  \emph{Can a single online forecaster achieve sublinear regret simultaneously for an infinite class of proper losses?}
\end{quote}
Moreover, the constant (in $T$) regret guarantee of~\cite{chernov2009prediction} for a finite class of mixable proper losses raises a further question:
\begin{quote}
  \emph{Are faster rates achievable for special infinite loss classes such as the class of all mixable proper losses?}
\end{quote}

\subsection{Our Results}
We give both negative and positive answers, revealing a separation between unrestricted proper losses, the full class of mixable proper losses, and finite collections of mixable proper losses, as well as separations between the contextual and the non-contextual setting.\footnote{The non-contextual setting is a special case of our problem where the benchmark class $\calF$ is the set of all constant mappings, making the contexts meaningless. \label{fn:non-contextual}}
More specifically, our contributions are as follows (throughout, we assume a finite benchmark class $\calF$ for simplicity, but our results generalize directly to a general $\calF$ as long as an oracle for online learning over $\calF$ is given, in the same vein as~\cite{okoroafor2025near, farina2026blackbox}).

\paragraph{Impossibility for all bounded proper losses.}
Our first result is a negative answer to the first question, showing that even when $\calF$ consists of only two linear predictors, no forecaster can achieve sublinear regret simultaneously for all bounded proper losses.

\begin{restatable}{theorem}{impossibility}
  \label{thm:impossibility}
  There exists a fixed benchmark class $\calF \subseteq [0,1]^{\calX}$ of two affine predictors for $\calX=[0,3/4]$ such that, for every online forecaster and every horizon $T$, there exist a sequence of context-outcome pairs $(x_1, y_1), \ldots, (x_T, y_T)$ and a proper loss with range $[0,1]$ such that $\E\sbr{\Reg_{\ell, \calF}(T)} \geq \frac{T}{12}$.
\end{restatable}

Thus, the obstacle to extending the aforementioned finite-basis arguments to this class is fundamental, rather than merely a limitation of existing analysis.
We also note that this impossibility result is in stark contrast to the non-contextual case where $\order(\sqrt{T})$ regret simultaneously for all bounded proper losses is achievable~\citep{kleinberg2023u, luo2024optimal}.
Finally, as pointed out by~\cite{okoroafor2025near}, every bounded proper loss has bounded variation, and thus \pref{thm:impossibility} also implies that no sublinear omniprediction regret guarantee is possible for the class of all losses with bounded variation.

\paragraph{Optimal omniprediction for all mixable proper losses.}
Despite the negative result to the first question, we next provide a positive answer to the second question.

\begin{restatable}{theorem}{mixable}
  \label{thm:mixable}
  There exists an online forecaster (\pref{alg:mixable}) such that for every mixable proper loss $\ell$,
  \begin{equation}
    \Reg_{\ell, \calF}(T)
    =
    \otil\rbr{\frac{T^{1/3}(\log|\calF|)^{2/3}}{\beta_\ell^\star}
    }
    \label{eq:universal-mixable}
  \end{equation}
  where $\otil(\cdot)$ suppresses dependence on $\log T$ factors.
\end{restatable}

Thus, while omniprediction with respect to all proper losses is impossible, as long as we restrict our attention to mixable losses, it not only becomes possible again but also admits faster (than $\order(\sqrt{T})$) rates.
On the other hand, compared to the finite class result of~\citep{chernov2009prediction} where constant regret is achievable, our regret bound is of order $T^{1/3}$.
Somewhat surprisingly, we further show that this rate is in fact optimal, even if we only consider the subclass of all Lipschitz proper losses.

\begin{restatable}{theorem}{mixableLB}
  \label{thm:lower-bound-mixable}
There exists a fixed benchmark class
$\calF\subseteq[0,1]^{\calX}$ of two linear predictors
for $\calX=[0,1]^2$ such that, for every online forecaster, every $\beta > 0$, and every $T$, there exist a sequence of context-outcome pairs $(x_1, y_1), \ldots, (x_T, y_T)$ and a $\frac{2}{\beta}$-Lipschitz (hence also $\beta$-mixable) bounded proper loss such that $\E\sbr{\Reg_{\ell, \calF}(T)} \geq \Omega\rbr{\frac{T^{1/3}}{\beta}}$.
\end{restatable}

This shows a separation between learning over the full class of mixable proper losses and a finite subset of it.
Moreover, it in fact also shows another separation between the contextual and the non-contextual case: in the non-contextual setting, \cite{luo2024optimal} showed that $\order(L\log T)$ regret is achievable simultaneously for all $L$-Lipschitz proper losses, while \pref{thm:mixable} and \pref{thm:lower-bound-mixable} together show that $LT^{1/3}$ is the best achievable rate for this loss class.

\paragraph{A best-of-all-worlds forecaster.}
Finally, in light of different possible omniprediction guarantees from both the literature and our work, we propose one single algorithm that achieves \textit{all} of them simultaneously.

\begin{restatable}{theorem}{unified}
  \label{thm:unified}
  For any fixed finite class $\calLbd$ of losses with bounded range $[0,1]$ and any fixed finite class of mixable proper losses $\calLmix$,
  there exists an online forecaster (\pref{alg:unified}) that guarantees all of the following simultaneously:
  \begin{enumerate}
    \item for every $\ell \in\calLmix$,
      \begin{equation}
        \Reg_{\ell,\calF}(T)
        =
        \order\rbr{\frac{\log\rbr{|\calF||\calLmix|}}{\beta_{\ell}^\star}}
        ;
        \label{eq:unified-listed-mixable}
      \end{equation}

    \item for every mixable proper loss $\ell$ (whether or not it belongs to $\calLmix$),
      \begin{equation}
        \Reg_{\ell,\calF}(T)
        =\otil\rbr{\frac{T^{1/3}(\log|\calF|)^{2/3}}{\beta_\ell^\star}}
        ;
        \label{eq:unified-universal-mixable}
      \end{equation}

    \item for every $\ell\in\calLbd$,
      \begin{equation}
        \Reg_{\ell,\calF}(T)
        =
        \order
        \rbr{
          \sqrt{T\log\rbr{|\calF||\calLbd|}}
        }.
        \label{eq:unified-bounded}
      \end{equation}
  \end{enumerate}
\end{restatable}

\begin{table}[t]
  \caption{Comparisons of omniprediction regret bounds on $\Reg_{\ell,\calF}(T), \forall \ell \in \calL$ for different loss families $\calL$. Benchmark class $\calF$ is assumed to be finite for simplicity. }
  \label{tab:summary}
  \vspace{5pt}
  \centering
  \renewcommand{\arraystretch}{2}
  \resizebox{\linewidth}{!}{%
    \begin{tabular}{c|c|c}
      \toprule
      \textbf{Loss family $\calL$} & \textbf{Prior results} & \textbf{Our results} \\
      \midrule
      \makecell{Finite class of \\ mixable proper losses}
      & \makecell{$\order\rbr{\frac{\log(|\calF||\calL|)}{\beta_\ell^\star}}$ \\ \citep{chernov2009prediction}}
      & \makecell{Same \\ \pref{alg:unified} and \pref{thm:unified}} \\

      \hline

      \makecell{Finite class of \\ bounded losses}
      & \makecell{$\order\rbr{\sqrt{T\log(|\calF||\calL|)}}$ \\ \citep{okoroafor2025near, farina2026blackbox}}
      & \makecell{Same \\ \pref{alg:unified} and \pref{thm:unified}} \\

      \hline

      All mixable proper losses
      & None
      & \makecell{$\otil\rbr{\frac{T^{1/3}(\log|\calF|)^{2/3}}{\beta_\ell^\star}}$ \\
        \pref{alg:mixable} and \pref{thm:mixable} \\
        \pref{alg:unified} and \pref{thm:unified} \\
        matching lower bound \pref{thm:lower-bound-mixable}
      } \\

      \hline

      \makecell{All bounded proper losses or \\ all bounded variation losses}
      & \makecell{$\otil\rbr{\sqrt{T\log(|\calF||\Gamma|)}}$
        if $\calF$ is $\Gamma$-valued \\ for a finite $\Gamma$ \citep{okoroafor2025near}
      }
      & \makecell{$\Omega(T)$ \\ \pref{thm:impossibility}} \\

      \hline

      \makecell{All $L$-Lipschitz losses}
      & \makecell{$\otil(L\sqrt{T\log|\calF|})$ \\
      \citep{okoroafor2025near}}
      & \makecell{Same \\ \pref{alg:unified} and \pref{cor:unified}} \\

      \hline

      \makecell{All generalized linear \\ model (GLM) losses}
      & \makecell{$\otil(\sqrt{T\log|\calF|})$ \\
      \citep{okoroafor2025near}}
      & \makecell{Same \\ \pref{alg:unified} and \pref{cor:unified}} \\
      \bottomrule
    \end{tabular}%
  }
\end{table}

By including squared loss and log loss in the finite mixable family $\calLmix$ and including suitable finite bases in the finite bounded family $\calLbd$, we further obtain a single forecaster with the following strong omniprediction guarantees.

\begin{restatable}{corollary}{unifiedCor}
  \label{cor:unified}
  There exists an online forecaster that guarantees all of the following simultaneously:
  \begin{enumerate}
    \item for both squared loss $\ell(p,y) = (p-y)^2$ and log loss $\ell(p,y) = -y\log p - (1-y)\log(1-p)$,
      \begin{equation*}
        \Reg_{\ell,\calF}(T)
        =
        \order\rbr{\log|\calF|}
        ;
      \end{equation*}

    \item for every mixable proper loss $\ell$,
      \begin{equation*}
        \Reg_{\ell,\calF}(T)
        =\otil\rbr{\frac{T^{1/3}(\log|\calF|)^{2/3}}{\beta_\ell^\star}}
        ;
      \end{equation*}


    \item for every $L$-Lipschitz loss $\ell$ (that is not necessarily proper),
      \begin{equation*}
        \Reg_{\ell,\calF}(T)
        =
        \otil
        \rbr{L
          \sqrt{T\log|\calF|}
        };
      \end{equation*}

    \item for every Generalized Linear Model (GLM) loss $\ell$,
      \begin{equation*}
        \Reg_{\ell,\calF}(T)
        =
        \otil
        \rbr{
          \sqrt{T\log|\calF|}
        };
      \end{equation*}

    \item for the $0$-$1$ loss $\ell(p,y)=(1-y)\Ind{p\ge1/2}+y\Ind{p<1/2}$,
      \begin{equation*}
        \Reg_{\ell,\calF}(T)
        =
        \otil
        \rbr{
          \sqrt{T\log|\calF|}
        }.
      \end{equation*}
  \end{enumerate}
\end{restatable}

Table~\ref{tab:summary} summarizes all our results and the comparisons with the closest prior work.

\subsection{Technical Overview}\label{sec:techincal_overview}

\paragraph{Impossibility for all bounded proper losses.}
The hardness stems from the fact that bounded proper losses can have discontinuities at arbitrary thresholds.
Indeed, it is well-known in the online learning literature that one-dimensional threshold functions are not online learnable; see e.g.~\cite[Theorem~8.2]{rakhlin2014statistical}.
The proof is based on an infinite binary tree that keeps cutting an interval at a node into two equal halves as its two children.
For a sequence of i.i.d.~binary outcomes, the contexts are then generated according to the path of the tree that is determined by these binary outcomes.
The difficulty of learning threshold functions boils down to the fact that no matter what the binary outcome sequence is, there always exists a threshold function that correctly classifies all examples.

In our problem, on the other hand, we are only competing with a finite benchmark class, so for a fixed loss, we cannot always find a perfect benchmark any more.
However, at the same time we are not constrained to a fixed loss function in omniprediction.
Indeed, we can effectively simulate infinite threshold classifiers via picking different proper losses since, as mentioned, proper losses can have discontinuities at arbitrary thresholds.
This is the key idea of our proof, where we only need two simple affine benchmarks $f_1(x)=x$ and $f_2(x)=x+1/4$, corresponding to two of the aforementioned binary trees with every node being $1/4$ apart from each other, so that no matter what the i.i.d.~binary outcomes turn out to be, one of the benchmarks must suffer zero loss on a particular threshold-based proper loss functions, while the learner must suffer $\Omega(T)$ loss in expectation.
The details can be found in \pref{app:impossibility}.

\paragraph{$\Omega(T^{1/3})$ lower bound for Lipschitz proper losses.}
To establish the $\Omega(T^{1/3})$ lower bound for Lipschitz losses, we divide the horizon into $K$ phases, each with a distinct Bernoulli mean $v_j$. We consider two fixed linear predictors, $f_+$ and $f_-$, which predict slightly above and below $v_j$, respectively, so that $f_+$ is favored by an excess of outcome $1$, while $f_-$ is favored by an excess of outcome $0$. These random fluctuations are captured by $S_j=\sum_{t\in\mathcal I_j}(y_t-v_j)$ where $\mathcal I_j$ denotes the rounds in the $j$-th phase.

Note that for a fixed proper loss, truthfully reporting $v_j$ in each phase $j$ is clearly optimal in expectation due to properness.
Therefore, the key is to decide the loss function after the realization of the outcomes.
More specifically, we construct a fixed family of Lipschitz proper losses whose curvature can be independently adjusted 
on each disjoint interval $[v_j-h_j,v_j+h_j]$ for $h_j=\Theta(\sqrt{K/T})$. After observing all outcomes, we select two losses that place additional curvature on intervals with positive and negative $S_j$, favoring $f_+$ and $f_-$, respectively. Crucially, the sum of these two losses is independent of the outcomes, which means the optimal strategy for the learner in terms of minimizing the expected sum of these two losses is still to  truthfully report $v_j$ in each phase $j$.
Meanwhile, the two benchmark comparisons capture fluctuations of both signs, yielding an expected advantage of $\Omega(1)$ per phase and thus $\Omega(K)$ overall regret.
Finally, to make sure that the intervals $[v_j-h_j,v_j+h_j]$ are disjoint, the largest possible value of $K$ is of order $T^{1/3}$, which leads to the $\Omega(T^{1/3})$ lower bound.
See \pref{app:lower} for details.

\subsubsection{Key ideas behind algorithms and upper bounds}\label{sec:algorithm_overview}
For upper bounds, our approach builds on the framework of~\cite{farina2026blackbox}, which we briefly review below using our notations and in a way that is slightly different from (but equivalent to) their original presentation.
The starting point is the following regret identity:
for any loss function $\ell$ and benchmark $f\in\calF$, the regret of the learner against $f$ can be written as
\begin{align}
  \ell(\pi_\ell(P_t), y_t) - \ell(f(x_t), y_t)
  &= b_\ell(P_t, f(x_t))(y_t - P_t) - d_\ell(P_t, f(x_t)) \label{eq:regret_identity} \\
  &\leq b_\ell(P_t, f(x_t))(y_t - P_t) \label{eq:regret_identity_dropping_d}
\end{align}
for some loss-dependent functions $b_\ell$ and $d_\ell$;
their concrete definitions are deferred to \pref{sec:prelim}, but one should treat $b_\ell$ as a certain test function (using the terminology of~\cite{farina2026blackbox}) and $d_\ell$ as a properness gain term that is always nonnegative, explaining the inequality \pref{eq:regret_identity_dropping_d}.
Now, assume $\calL$ is finite and treat each benchmark-loss pair $(f, \ell)$ as an expert with reward $b_\ell(P_t, f(x_t))(y_t - P_t)$ in round $t$.
We can then use a standard expert algorithm, such as the classic multiplicative/exponential weight update of~\cite{freund1997decision}, to produce a distribution $W_{t}$ over all $|\calF||\calL|$ benchmark-loss pairs and ensure for each $(f, \ell)$,
\begin{equation}\label{eq:expert_guarantee}
  \sum_{t=1}^T b_\ell(P_t, f(x_t))(y_t - P_t) - \sum_{t=1}^T \E_{(f', \ell') \sim W_t}\sbr{b_{\ell'}(P_t, f'(x_t))(y_t - P_t)}
  \leq \order\rbr{\sqrt{T\log(|\calF||\calL|)}}.
\end{equation}
Therefore, if in each round $t$, we can make sure that, no matter what the outcome $y_t$ turns out to be, the forecast distribution $D_t$ always ensures $\E_{P_t\sim D_t}\sbr{G_t(P_t)(y_t-P_t)} \approx 0$ for an expected test function $G_t(P) = \E_{(f', \ell') \sim W_t}\sbr{b_{\ell'}(P, f'(x_t))}$, then combining \pref{eq:regret_identity_dropping_d} and \pref{eq:expert_guarantee} proves $\Reg_{\ell, \calF} = \order\rbr{\sqrt{T\log(|\calF||\calL|)}}$ for all $\ell \in \calL$.
Indeed, the aforementioned requirement on $D_t$ exactly corresponds to an \textit{Expected Variational Inequality} (EVI), which can be solved efficiently in time $\log(1/\epsilon)$ to make sure $\E_{P_t\sim D_t}\sbr{G_t(P_t)(y_t-P_t)} \leq \epsilon$ for any precision $\epsilon > 0$~\citep{zhang2025expected}.

\paragraph{V-shaped losses and continuous exponential weights.}
To apply the same idea to an infinite class of proper losses, we utilize the fact that, up to a prediction-independent term, every proper loss $\ell(p,y)$ of our interest can be expressed as a conic combination of \textit{V-shaped losses}: $\int_{0}^1 w_\ell(v)\lambda_v(p,y) dv$ where $\lambda_v$ is the V-shaped loss at threshold $v$ and $w_\ell(v)$ is the corresponding nonnegative weight (\pref{lem:automatic-curvature}).
It is thus natural to apply the framework of~\cite{farina2026blackbox} to the infinite class of V-shaped losses by learning the distribution $W_t$ via the continuous version of exponential weights (see e.g.~\citep{hoeven2018many}).

While continuous exponential weights does not guarantee a regret bound against any single V-shaped loss, 
it admits a regret bound against a \textit{mixture} of the V-shaped losses, in the form of $\frac{\KLComp}{\eta} + \eta T$, where $\eta$ is a learning rate and $\KLComp$ is a complexity term that depends on the KL divergence between the mixture weight and a prior distribution.
In our case, we naturally pick the mixture weight according to the normalized version of $w_\ell(v)$.
Bounding $\KLComp$ then requires special properties of the weight $w_\ell(v)$, which corresponds to special structures of the proper loss class, such as Lipschitzness or more generally mixability.

\paragraph{Warm-up for Lipschitz class and second-order correction.}
To best illustrate this idea, we use the Lipschitz proper loss class as a warm-up in \pref{sec:all-lipschitz}.
It can be shown that every $L$-Lipschitz proper loss ensures $w_\ell(v) \leq 2L$ and thus $\KLComp$ can be well bounded; see \pref{lem:lipschitz-comparison}.

However, the discussion so far (and specifically the bound of the form $\frac{\KLComp}{\eta} + \eta T$) only leads to a regret bound of order $\sqrt{T}$.
To further improve it to $T^{1/3}$, we need the following two ideas:
first, incorporate a second-order correction term to the exponential weight updates that leads to a second-order regret bound of the form
\begin{equation}\label{eq:second_order_regret}
  \frac{\KLComp}{\eta} + \sum_{t=1}^T \int_0^1 w_\ell(v) \eta b_\ell(P_t, f(x_t))^2 dv,
\end{equation}
an idea that has been explored in the literature~\citep{hazan2010extracting, steinhardt2014adaptivity, wintenberger2017optimal, chen2021impossible} (\pref{lem:lipschitz-comparison});
second, combine the second-order term above and the important negative term $-d_\ell(P_t, f(x_t))$ in \pref{eq:regret_identity} (that was dropped in earlier analysis) and show that together they are of order $\order(\eta^2)$ instead of $\order(\eta)$ (\pref{lem:lipschitz-cancellation}):
\begin{equation}\label{eq:cancellation}
  \int_0^1 w_\ell(v) \rbr{\eta b_\ell(P_t, f(x_t))^2 - d_\ell(P_t, f(x_t))} dv \lesssim  L \eta^2.
\end{equation}
Picking $\eta$ appropriately finally leads to a regret bound of order $T^{1/3}$.

\paragraph{Extension to mixable class.}
To handle the more general class of mixable proper losses, several more ideas are required (\pref{sec:all-mixable}).
The difficulty comes from the fact that for each $\beta$-mixable proper loss $\ell$, the corresponding weight $w_\ell(v)$ can be as large as $\frac{1}{\beta v(1-v)}$ (hence no longer always bounded; see \pref{lem:automatic-curvature}).
To compensate this, we naturally run continuous exponential weights with a prior that is proportional to $\frac{1}{v(1-v)}$ for the V-shaped loss parameterized by threshold $v$, so that $\KLComp$ is again bounded (\pref{lem:mixable-comparison}).
Noticing that $\int_0^1 \frac{dv}{v(1-v)}$ (the normalization factor for the prior) is unbounded, we also naturally only consider $v$ from the interval $[\delta, 1-\delta]$ for some small $\delta$ of order $1/T$.

Moreover, due to the potentially unbounded $w_\ell(v)$, \pref{eq:cancellation} no longer holds.
To address this issue, we come up with a new second-order correction term to be incorporated into the continuous exponential weight update.
Together with a new expected test function $G_t$ designed accordingly,
this 
leads to the following new second-order regret bound that, compared to \pref{eq:second_order_regret}, contains an extra \textit{Bernoulli variance factor} $P_t(1-P_t)$ (\pref{lem:mixable-comparison}):
\begin{equation}\label{eq:second_order_regret_refined}
  \frac{\KLComp}{\eta} + \sum_{t=1}^T \int_0^1 w_\ell(v) \eta b_\ell(P_t, f(x_t))^2 \underbrace{P_t(1-P_t)}_{\text{extra factor}} dv.
\end{equation}
When combined with the negative term $-d_\ell(P_t, f(x_t))$,
this extra Bernoulli variance factor exactly helps cancel the potentially large $w_\ell(v)$ (recall $w_\ell(v) \leq \frac{1}{\beta v(1-v)}$), leading to the following analogue of \pref{eq:cancellation} (\pref{lem:mixable-cancellation}):
\begin{equation}\label{eq:cancellation_refined}
  \int_0^1 w_\ell(v) \rbr{\eta b_\ell(P_t, f(x_t))^2 P_t(1-P_t) - d_\ell(P_t, f(x_t))} dv \lesssim  \frac{\eta^2}{\beta}.
\end{equation}
Picking $\eta$ appropriately again leads to the claimed regret bound of order $T^{1/3}$.

\paragraph{Combining everything.}
Finally, to achieve the best-of-all-worlds guarantee of \pref{thm:unified}, it suffices to combine the algorithmic components for both the finite class result and the infinite class result under the same framework of~\cite{farina2026blackbox}; see \pref{sec:unified}.

\subsection{Related Work}
\paragraph{Omniprediction and multicalibration.}
Omniprediction in the batch setting was introduced by~\cite{gopalan2022omnipredictors}, who showed how a single forecast can be post-processed to compete across a family of losses.
They achieved so by drawing a connection to multicalibration~\citep{hebert2018multicalibration}.
Motivated by the role of swap regret in online learning,
\cite{gopalan2023swap} introduced the stronger ``swap version" of both omniprediction and multicalibration.
\cite{garg2024oracle} further extended these concepts to the online setting, which is the focus of our work.
We point out that, however, standard online-to-batch reductions can be used to convert online regret bounds to sample complexity bounds in the batch setting; see e.g.~\cite{okoroafor2025near, luo2025improved}.

\cite{okoroafor2025near} are the first to develop near-optimal omniprediction guarantees for several loss families.
More recently, \cite{farina2026blackbox} developed a different black-box approach that achieves multicalibration/omniprediction via standard online learning.
As mentioned, our methods are built on this approach.

In a concurrent work, \cite{okoroafor2026swap} studied the stronger swap version of omniprediction and show that $\otil(\sqrt{T\log(|\calF||\Gamma|)})$ swap regret is achievable simultaneously for all bounded proper losses if $\calF$ is $\Gamma$-valued for a finite set $\Gamma$, similar to the non-swap version of~\cite{okoroafor2025near}.
Our impossibility result (\pref{thm:impossibility}) shows that the restriction to a finite $\Gamma$ is unavoidable.
On the other hand, \cite{okoroafor2026swap} also showed that $\otil(T^{1/3}(\log|\calF|)^{2/3})$ swap regret is achievable for all convex and $1$-Lipschitz proper losses.
This result shares some similarity with our \pref{thm:mixable}, except that their regret notion is stronger while our loss class is broader (more concretely, we require neither convexity nor Lipschitzness but only mixability).
Their approach to achieve this fast rate is via a certain second-order form of multicalibration.

\paragraph{Non-contextual setting and U-calibration.}
As mentioned in \pref{fn:non-contextual}, the non-contextual setting is a special case where $\calF$ contains all constant mappings.
Omniprediction in this setting is referred to as U-calibration~\citep{kleinberg2023u}.
As discussed earlier, our work shows several gaps between U-calibration and omniprediction, with many stronger guarantees achievable for the former but not the latter.

Our best-of-all-worlds result (\pref{thm:unified}) is in spirit similar to the result of~\cite{frongillo2026toward} and~\cite{dewasurendra2026dirichlet} for U-calibration.
Specifically, they developed a single U-calibrated forecaster with $\otil(\sqrt{T})$ regret simultaneously for all bounded proper losses and also $\order(\log T)$ regret for smooth proper losses.
Their approaches are quite different from and much simpler than ours, due to the lack of contexts.

\paragraph{Multiclass omniprediction.}
Our results are restricted to binary outcomes.
While the finite-class results can be generalized to the multiclass case according to~\cite{farina2026blackbox}, it is unclear how to do so for the infinite-class result since there is no similar V-shaped basis for the multiclass case~\citep{kleinberg2023u}.
Moreover, the sample complexity for multiclass omniprediction is often exponential in the number of classes; see e.g.~\cite{hu2026multiclass}.

\renewcommand{\theHtheorem}{\thesection.\arabic{theorem}}
\renewcommand{\theHlemma}{\theHtheorem}
\renewcommand{\theHcorollary}{\theHtheorem}
\renewcommand{\theHproposition}{\theHtheorem}
\renewcommand{\theHremark}{\theHtheorem}

\section{Preliminaries}
\label{sec:prelim}

\textbf{Notation.}
For every positive integer $n$, $[n]$ denotes the set $\{1,\ldots,n\}$. We write $[x]_+=\max\{x,0\}$. The distributions $\Unif(S)$, $\Ber(p)$, and $\delta_x$ are uniform on a finite nonempty set $S$, Bernoulli with mean $p$, and Dirac delta at $x$, respectively. The product measure $P\otimes Q$ describes independent draws from $P$ and $Q$. For any probability measures $Q,P$ on the same measurable space, $Q\ll P$ means $P(A)=0$ implies $Q(A)=0$ for every measurable $A$. With $\dmeas Q/\dmeas P$ denoting the density of $Q$ relative to $P$, $\KL(Q\|P)=\int\log(\dmeas Q/\dmeas P)\dmeas Q$ when $Q\ll P$, and $+\infty$ otherwise. We also write $\E_Q[g]=\int g\dmeas Q$. The measure $\Leb$ is Lebesgue measure, with respect to which almost-everywhere statements on intervals are understood. The clipping operator is $\ClipOp(x,[a,b])=\min\{b,\max\{a,x\}\}$.

\paragraph{Proper losses and two subclasses.}
We recall that a loss $\ell$ is proper if
    $
    \E_{Y\sim\Ber(p)}\sbr{\ell(p,Y)}
    \le
    \E_{Y\sim\Ber(p)}\sbr{\ell(q,Y)},
    \forall p,q\in[0,1],
    $
    that is, predicting $p$ is optimal when the outcome is indeed drawn from $\Ber(p)$.
    In this case we let $\pi_\ell(p)$, the post processing function $\argmin_{q\in[0,1]} \E_{Y\sim\Ber(p)}\sbr{\ell(q,Y)}$, be simply $p$.

We consider two subclasses of proper losses: $\LipLosses{L}$ and $\MixLosses{\beta}$, where the first one represents the class of all $L$-Lipschitz proper losses and the second represents the class of all $\beta$-mixable proper losses.
Specifically, a loss $\ell$ is $L$-Lipschitz if its partial loss function $\ell(\cdot, y)$ is $L$-Lipschitz for each $y \in \{0,1\}$, that is $|\ell(p,y)-\ell(q,y)|\le L|p-q|$ for all $p,q\in[0,1]$.
As mentioned in \pref{sec:intro}, a loss $\ell$ is $\beta$-mixable if, for every distribution $D$ over $[0,1]$, there exists $p \in [0,1]$ such that
    $
    \exp\rbr{-\beta\ell(p,y)}
    \ge
    \E_{P\sim D}\exp\rbr{-\beta\ell(P,y)},
    \forall y\in\cbr{0,1}.
    $
    We recall that $\beta_\ell^\star$ is largest constant $\beta$ such that $\ell$ is $\beta$-mixable.

\paragraph{Threshold representation and curvature bounds.}
Up to a prediction-independent term, every Lipschitz/mixable proper loss $\ell$ can be written as a conic combination of V-shaped losses.
Specifically, for any $v\in [0,1]$, the V-shaped loss $\lambda_v$ parameterized by threshold $v$ is defined through
\begin{equation}\label{eq:v-shaped-loss}
\lambda_v(q,1)=(1-v)\Ind{q<v} \quad\text{and}\quad \lambda_v(q,0)=v\Ind{q\ge v}.\footnote{This is called V-shaped loss because its Bayes risk $\E_{Y\sim\Ber(p)}\sbr{\lambda_v(p,Y)}$ is the V-shaped function $\min\{(1-v)p, v(1-p)\}$.
We also note that our definition is slightly different from but equivalent to (up to shifting and scaling) the definition of prior work such as~\cite{kleinberg2023u, okoroafor2025near}.}
\end{equation}
Then, the following lemma holds (see \pref{app:prelim} for the proof).

\begin{lemma}
\label{lem:automatic-curvature}
If a proper loss $\ell$ is Lipschitz or mixable, then it can be written as
\begin{equation}\label{eq:v-shaped-rep}
  \ell(p,y)
  =C_\ell(y)+\int_0^1w_\ell(v)\lambda_v(p,y)\dmeas v
\end{equation}
for all $p\in[0,1]$ and $y\in\{0,1\}$, where $C_\ell(y)$
is independent of the prediction and $w_\ell$ is nonnegative
and measurable.
Moreover, for any proper loss admitting the representation
\pref{eq:v-shaped-rep} and any $L,\beta>0$,
\begin{itemize}
\item $\ell\in\LipLosses{L}$ if and only if
$w_\ell(v)\le \frac{L}{\max\{v,1-v\}}$ almost everywhere;
\item $\ell\in\MixLosses{\beta}$ if and only if
$w_\ell(v)\le \frac{1}{\beta v(1-v)}$ almost everywhere.
\end{itemize}
Consequently, if $\ell\in\LipLosses{L}$, then $w_\ell(v) \leq 2L$ almost everywhere and also
$\ell\in\MixLosses{2/L}$.
\end{lemma}



\paragraph{Loss differences and test functions.}
For any loss $\ell$, write the loss difference between two outcomes at the same prediction $q$ as $\Delta_\ell(q)=\ell(q,1)-\ell(q,0)$. 
Further define 
\[
  b_\ell(p,q)=\Delta_\ell(\pi_\ell(p))-\Delta_\ell(q)
  \quad\text{and}\quad
  d_\ell(p,q)=\E_{Y\sim\Ber(p)}[\ell(q,Y)-\ell(\pi_\ell(p),Y)].
\]
Then, the regret identity of \pref{eq:regret_identity} can be verified as follows: for any $p, q\in [0,1]$ and $y\in\{0,1\}$,
\begin{align}
\ell(\pi_\ell(p), y) - \ell(q, y)
&= y b_\ell(p,q) + \ell(\pi_\ell(p), 0) - \ell(q, 0) \notag\\
&= (y-p)b_\ell(p,q) + p b_\ell(p,q) + \ell(\pi_\ell(p), 0) - \ell(q, 0) \notag\\
&= (y-p)b_\ell(p,q) + p\rbr{\ell(\pi_\ell(p), 1) - \ell(q, 1)} + (1-p)\rbr{\ell(\pi_\ell(p), 0) - \ell(q, 0)} \notag\\
&= (y-p)b_\ell(p,q) - d_\ell(p,q). \label{eq:regret_identity_general}
\end{align}
Following~\cite{farina2026blackbox}, we call $b_\ell(p,q)$ a test function.
Also note that by the definition of $\pi_\ell$, $d_\ell(p,q)$ is always nonnegative and can be treated as properness gain.



For a V-shaped loss $\lambda_v$, we use $b_v(p,q)$ and $d_v(p,q)$ as the shorthands for $b_{\lambda_v}(p,q)$ and $d_{\lambda_v}(p,q)$. It is straightforward to verify that they take the following simple forms:
\[
  b_v(p,q)=\Ind{p<v\le q}-\Ind{q<v\le p} \quad\text{and}\quad
  d_v(p,q)=|v-p|b_v(p,q)^2.
\]
Therefore, when $\ell$ is either Lipschitz or mixable,  \pref{eq:v-shaped-rep} and \pref{eq:regret_identity_general} together imply: for any $p, q\in [0,1]$ and $y\in\{0,1\}$,
\begin{equation}
  \ell(p,y)-\ell(q,y)
  =\int_0^1w_\ell(v)\rbr{b_v(p,q)(y-p)-d_v(p,q)}\dmeas v,
  \label{eq:local-v-representation}
\end{equation}
a regret identity that is critical to our analysis.

\paragraph{EVI on an interval.}
As discussed, a key component of the algorithm of~\cite{farina2026blackbox} is a solver for an Expected Variational Inequality (EVI), whose efficiency is guaranteed due to~\cite{zhang2025evi}.
For our purpose, we only need to solve an EVI over the interval $I_\delta=[\delta,1-\delta]$ for some $\delta \in [0, 1/2)$, which admits simple solutions. 
The details are summarized in the following lemma.

\begin{lemma}[Binary EVI on a closed interval]
  \label{lem:interior-evi}
  For every $0\le\delta<1/2$, every $G:I_\delta\to\mathbb R$ satisfying $1+G(p)(y-p)\ge0$ for all $p\in I_\delta$ and $y\in\{0,1\}$, and every $\epsilon>\delta/(1-\delta)$, there is a distribution $D$ on $I_\delta$ with at most three support points such that
  \[
    \E_{P\sim D}[G(P)(y-P)]\le\epsilon,
    \qquad\forall y\in \{0,1\}.
  \]
  When $\delta>0$, $D$ may be supported on any finite grid containing the endpoints of $I_\delta$ whose adjacent gaps are at most $\delta$.
\end{lemma}

\section{Algorithms and Results}
\label{sec:algorithms}

Following the overview in \pref{sec:algorithm_overview},
we start with a warm-up in \pref{sec:all-lipschitz} and consider omniprediction for the class of all Lipschitz proper losses.
We then extend the result to the class of all mixable proper losses in \pref{sec:all-mixable} and finally combine it with other finite-class guarantees in \pref{sec:unified}.

\subsection{Warm-up: all Lipschitz proper losses}
\label{sec:all-lipschitz}
As discussed in \pref{sec:algorithm_overview},
the V-shaped representation in \pref{eq:local-v-representation} suggests an expert for each predictor--threshold pair $(f,v)\in\calF\times[0,1]$, with reward $z_{t,f,v}=\E_{P\sim D_t}[b_v(P,f(x_t))(y_t-P)]$ under the round-$t$ forecast distribution $D_t$.  
We thus run (a variant of) continuous exponential weights over these infinitely many experts.
Specifically, the algorithm deploys a uniform prior $\pi_0=\operatorname{Unif}(\calF)\otimes\mu_0$ over the experts
where $\mu_0$ is the uniform probability measure on $[0,1]$,
and maintains a nonnegative weight density $W_{t,f}(v)$ updated via (for a learning rate $\eta>0$)
\begin{equation}\label{eq:W_update}
      W_{t+1,f}(v)=W_{t,f}(v)\exp\rbr{\eta z_{t,f,v}-\eta^2z_{t,f,v}^2 },
\end{equation}
starting from $W_{1,f}(v)=1$ for all $f\in\calF$ and $v\in[0,1]$.
Compared to the standard exponential weights, an extra second-order correction term $-\eta^2z_{t,f,v}^2$ is incorporated into the update, 
an idea that has been explored in the literature for the discrete version of exponential weights starting from~\cite{hazan2010extracting}.

The expected test function at time $t$ is then the following mixture of V-shaped test functions:
\begin{equation}
  G_t(p)=\frac{\eta}{\Phi_t}
  \E_{(f,v)\sim\pi_0}\sbr{W_{t,f}(v)b_v(p,f(x_t))}, 
  \label{eq:lipschitz-test}
\end{equation}
where $\Phi_t=\E_{(f,v) \sim \pi_0}[W_{t,f}(v)]$ is the total mass (the normalization factor).
We then solve the following EVI: find a forecast distribution $D_t$ supported on $[0,1]$ such that $\E_{P\sim D_t}[G_t(P)(y-P)]\le\epsilon_t$ for any $y\in\{0,1\}$ and precision $\epsilon_t = 1/T$.
The complete pseudocode is shown in \pref{alg:lipschitz}.

\begin{algorithm}[t]
  \caption{An online omnipredictor for all Lipschitz proper losses}
  \label{alg:lipschitz}
  \textbf{Input}: a horizon $T$, a finite prediction class $\calF$, a learning rate $0<\eta\le1/2$, and an EVI solver.  

  \textbf{Initialization:} prior $\pi_0=\operatorname{Unif}(\calF)\otimes\mu_0$, where $\mu_0$ is uniform on $[0,1]$; $\epsilon_t=1/T$, and $W_{1,f}(v)=1$.

  \For{$t=1,\ldots,T$}{
    Nature reveals $x_t$.

    Compute total mass $\Phi_t=\E_{(f,v) \sim \pi_0}[W_{t,f}(v)]$ and the expected test function 
    \[
    G_t(p)=\frac{\eta}{\Phi_t}\E_{(f,v)\sim\pi_0}\sbr{W_{t,f}(v)b_v(p,f(x_t))}.
    \]

    Use the EVI solver to find a distribution $D_t$ over $[0,1]$, 
    satisfying
    \[
      \E_{P\sim D_t}[G_t(P)(y-P)]\le\epsilon_t,
      \qquad\text{for all } y\in \{0,1\}.
    \]

    Nature reveals outcome $y_t$.

    For every expert $(f,v)\in\calF\times[0,1]$, define its reward as $z_{t,f,v}=\E_{P\sim D_t}[b_v(P,f(x_t))(y_t-P)]$.

    Update the weight via an exponential update with a second-order correction term:
    \[
      W_{t+1,f}(v)=W_{t,f}(v)\exp\rbr{\eta z_{t,f,v}-\eta^2z_{t,f,v}^2}.
    \]
  }
\end{algorithm}


\paragraph{Regret analysis.}
Consider an $L$-Lipschitz proper loss $\ell$.
For any benchmark $f\in\calF$, 
\pref{eq:local-v-representation} gives the regret decomposition
\begin{equation}\label{eq:regret_identity_v_shaped}
  \sum_{t=1}^T\E_{P\sim D_t}[\ell(P,y_t)-\ell(f(x_t),y_t)]=\sum_{t=1}^T\int_0^1w_\ell(v)z_{t,f,v}\dmeas v- \sum_{t=1}^T\E_{P\sim D_t}\int_0^1w_\ell(v)d_v(P,f(x_t))\dmeas v.
\end{equation}
The first term is the reward of the target mixture, while the second is the negative properness gain.  The following lemma bounds the first term using the regret of continuous exponential weights and the EVI guarantee.

\begin{lemma}
  \label{lem:lipschitz-comparison}
  If $\ell$ is proper and $L$-Lipschitz, then for 
  any $f\in\calF$, \pref{alg:lipschitz} satisfies
  \[
    \begin{aligned}
      \sum_{t=1}^T\int_0^1w_\ell(v)z_{t,f,v}\dmeas v
      \le{}\order\rbr{\frac{L\log(|\calF|)}{\eta}}+\eta\sum_{t=1}^T\E_{P\sim D_t}\int_0^1w_\ell(v)b_v(P,f(x_t))^2(y_t-P)^2\dmeas v.
    \end{aligned}
  \]
\end{lemma}
\begin{proof}
For any fixed $(f,v)$, the weight ratio can be bounded as
\begin{align}
  \frac{W_{t+1,f}(v)}{W_{t,f}(v)}
  &=\exp\rbr{\eta z_{t,f,v}
  -\eta^2z_{t,f,v}^2 } \le 1+\eta z_{t,f,v}, \notag
\end{align}
where the inequality uses $e^{x-x^2}\le1+x$
for $|x|\le1/2$, since $|b_v|\le1$ and $\eta\le1/2$.
Averaging over $\pi_0$ and using the definition of $G_t$ gives
\begin{align*}
  \Phi_{t+1}
  &\le\E_{(f,v)\sim\pi_0}
  \sbr{W_{t,f}(v)(1+\eta z_{t,f,v})}\\
  &=\Phi_t\rbr{1+ \E_{P\sim D_t}
  \sbr{G_t(P)(y_t-P)}}\\
  &\le(1+\epsilon_t)\Phi_t,
\end{align*}
where the last inequality follows from the EVI condition
in \pref{alg:lipschitz}.
Thus, $\Phi_1=1$ and $\epsilon_t=1/T$
imply $\Phi_{T+1}\le \rbr{1+\frac{1}{T}}^T \leq e$.
  Therefore, for any probability distribution $Q\ll\pi_0$, Jensen's inequality gives 
  \begin{equation}
    \E_Q\sbr{\log W_{T+1}}-\KL(Q\|\pi_0)
    =\E_Q\sbr{\log\rbr{\frac{W_{T+1}}{\dmeas Q}\dmeas \pi_0}}
    \le\log\E_{\pi_0}\sbr{W_{T+1}}
    =\log\Phi_{T+1}\le1.
    \label{eq:prior-comparison}
  \end{equation}  
  Plugging the definition of $W_{T+1}$, rearranging, and using $z_{t,f,v}^2 \leq \E_{P\sim D_t}\sbr{b_v(P,f(x_t))^2(y_t-P)^2}$ (by Jensen's inequality) gives
  \[
    \sum_t \E_{(f,v)\sim Q}\sbr{z_{t,f,v}}
    \le\frac{(\KL(Q\|\pi_0)+1)}{\eta}
    +\eta\sum_t\E_{P\sim D_t}\E_{(f,v)\sim Q}\sbr{b_v(P,f(x_t))^2(y_t-P)^2}.
  \]
  Since our objective is to bound $\sum_{t=1}^T\int_0^1w_\ell(v)z_{t,f,v}\dmeas v$,
  we naturally pick the target distribution as $Q=\delta_f\otimes\nu_\ell$, where $\nu_\ell(\dmeas v)=\frac{1}{m}w_\ell(v)\dmeas v$ and $m=\int_0^1w_\ell(v)\dmeas v$ is the normalization factor, leading to
  \[
    \sum_t\int_0^1w_\ell(v)z_{t,f,v}\dmeas v
    \le\frac{m(\KL(Q\|\pi_0)+1)}{\eta}
    +\eta\sum_t\E_{P\sim D_t}\int_0^1w_\ell(v)b_v(P,f(x_t))^2(y_t-P)^2\dmeas v.
  \]
  It thus remains to bound the KL term.  The uniform prior on $\calF$ and the curvature bound $w_\ell\le2L$ from \pref{lem:automatic-curvature} yield
  \[
    \begin{aligned}
      \KL(Q\|\pi_0)
      &=\log |\calF|+\KL(\nu_\ell\|\mu_0)
      =\log |\calF|+\int_0^1\frac{w_\ell(v)}m\log\frac{w_\ell(v)}m\dmeas v\le\log |\calF|+\log\frac{2L}{m}.
    \end{aligned}
  \]
  Consequently, we have $m(\KL(Q\|\pi_0)+1)=\order(L\log(|\calF|))$, using $m\le2L$ and $s\log(1/s)\le1/e$ for $s=m/(2L)$. This proves the lemma.
\end{proof}


The next lemma then shows that the second-order term from the bound of \pref{lem:lipschitz-comparison} that is of order $\eta $ per round can be combined with the negative properness gain from \pref{eq:regret_identity_v_shaped} to lead to a smaller term of order $\eta^2$, a key to derive the final order $T^{1/3}$ regret bound.

\begin{lemma}
  \label{lem:lipschitz-cancellation}
  For any $L$-Lipschitz proper loss $\ell$,
  $0<\eta\le1/2$, and $p,q\in[0,1]$, we have
  \[
    \int_0^1w_\ell(v)\rbr{\eta b_v(p,q)^2-d_v(p,q)}\dmeas v
    \le L \eta^2.
  \]
\end{lemma}
\begin{proof}
  Using the definition $d_v(p,q)=|v-p|b_v(p,q)^2$ and the condition $w_\ell(v) \in [0, 2L]$, we have
  \begin{align}
   \int_0^1w_\ell(v)\rbr{\eta b_v(p,q)^2-d_v(p,q)}\dmeas v
   &\leq\int_0^1w_\ell(v)b_v(p,q)^2[\eta-|v-p|]_+\dmeas v \notag\\
   &\leq 2L \int_0^1 b_v(p,q)^2 [\eta-|v-p|]_+\dmeas v \notag \\
   &\leq 2L \int_0^\eta(\eta-s)\dmeas s= L\eta^2, \label{eq:lipschitz-cancellation}
  \end{align}
  where the last inequality is because $b_v(p,q)^2 \leq 1$ and the integrand $b_v(p,q)^2 [\eta-|v-p|]_+$ is non-zero only when $v$ is at most $\eta$ from $p$ and also in between $p$ and $q$ (otherwise $b_v(p,q)=0$).
\end{proof}

Combining these two lemmas and picking the optimal $\eta$ immediately leads to the following final regret guarantee for \pref{alg:lipschitz}.
We emphasize that the value of $\eta$ is \textit{independent of $L$} and the regret bound holds for all Lipschitz proper losses simultaneously.

\begin{theorem}
  \label{thm:universal-lipschitz}
  With $\eta=\min\cbr{\frac12,
    \rbr{\frac{\log |\calF|}{T}}^{1/3}}$,
    \pref{alg:lipschitz} ensures that for any proper and $L$-Lipschitz loss $\ell$,
  \[
    \Reg_{\ell,\calF}(T)
    =\order\rbr{L\rbr{T^{1/3}(\log|\calF|)^{2/3}+\log|\calF|}}.
  \]
\end{theorem}

\subsection{All mixable proper losses}
\label{sec:all-mixable}

We now extend the idea in \pref{sec:all-lipschitz} to handle all mixable proper losses.  The construction still uses exponential weights over predictor--threshold pairs and an EVI to select forecasts.  However, the curvature of a mixable loss can grow near the endpoints, which creates two problems.

The first problem is the choice of prior.  For every $\beta>0$ and $\ell\in\MixLosses{\beta}$, the curvature bound $w_\ell(v)\le1/\rbr{\beta v(1-v)}$ from \pref{lem:automatic-curvature} suggests weighting thresholds by $1/\rbr{v(1-v)}$.  This density has infinite mass on $[0,1]$, so it cannot be normalized into a prior.  To address this problem, we restrict thresholds and forecasts to $I_\delta=[\delta,1-\delta]$, with a small $\delta=1/\rbr{2(T+1)}$ that has a negligible effect on the regret.  The density is then integrable, with normalizing constant $M_\delta=\int_{I_\delta}[v(1-v)]^{-1}\dmeas v=2\log(2T+1)$.
This gives the threshold and expert priors
\begin{equation}\label{eq:prior_delta}
  \mu_\delta(\dmeas v)=\frac{\dmeas v}{M_\delta v(1-v)},\qquad
  \pi_\delta=\operatorname{Unif}(\calF)\otimes\mu_\delta.
\end{equation}

With the prior in place, the second problem is how to control the second-order term in the regret.  
As discussed in \pref{sec:techincal_overview},
the curvature bound $w_\ell(v)\le1/\rbr{\beta v(1-v)}$ that grows near the boundary is again the key issue,
and addressing it requires introducing an extra factor in the second-order term of the regret to offset the curvature.
Here, having shown the proof for how to cancel the second-order term with the properness gain in the Lipschitz case (\pref{lem:lipschitz-cancellation}),
we provide a more detailed explanation on this issue and its solution.
Specifically, in the proof for \pref{lem:lipschitz-cancellation} (in particular, \pref{eq:lipschitz-cancellation}),
we see that for a fixed $p$, the cost comes from all thresholds $v$ with $|v-p|<\eta$. 
For mixable losses, however, the same neighborhood can carry much more curvature when $p$ and thus $v$ are close to the boundary.  
If we could somehow restrict the region to $|v-p|<\eta p(1-p)$ instead,
then it shrinks as $p$ approaches the boundary as well, exactly offsetting the growth in curvature.

With a closer examination of the second-order regret term from \pref{lem:lipschitz-comparison}, we see that in order to obtain the extra $p(1-p)$ factor mentioned above, we need to replace 
the realized square $(y_t-p)^2$ in the second-order term by the Bernoulli variance $p(1-p)$ (as if $y_t$ was actually drawn from $p$).
To do so, we accordingly change the update rule of the weight function from \pref{eq:W_update} to the following (note the change from $z_{t,f,v}^2=\E_{P\sim D_t}[b_v(P,f(x_t))(y_t-P)]^2$ to $\E_{P\sim D_t}[P(1-P)b_v(P,f(x_t))^2]$):
\begin{equation}\label{eq:W_update_new}
W_{t+1,f}(v)=W_{t,f}(v)\exp\rbr{\eta z_{t,f,v}-\eta^2\E_{P\sim D_t}[P(1-P)b_v(P,f(x_t))^2]}.
\end{equation}
For the regret analysis to still go through in the same way as \pref{lem:lipschitz-comparison},
all we need to ensure is that the potential increase constraint $\Phi_{t+1} \leq (1+\epsilon_t)\Phi_t$ still holds (where $\Phi_t$ is as before the total mass $\E_{(f,v)\sim\pi_\delta}[W_{t,f}(v)]$).
To this end, we follow the previous analysis and first bound the ratio $\frac{W_{t+1,f}(v)}{W_{t,f}(v)}$ as (using Jensen's inequality):
\begin{align}
\frac{W_{t+1,f}(v)}{W_{t,f}(v)} 
&= \exp\rbr{\eta \E_{P\sim D_t}[b_v(P,f(x_t))(y_t-P)]-\eta^2\E_{P\sim D_t}[P(1-P)b_v(P,f(x_t))^2]} \notag\\
&\leq \E_{P\sim D_t}\sbr{\exp\rbr{\eta b_v(P,f(x_t))(y_t-P)-\eta^2 P(1-P)b_v(P,f(x_t))^2}}. \label{eq:W_ratio_Jensen2}
\end{align}
Here, with the new second-order term, we can no longer apply the inequality $e^{x-x^2}\leq1+x$ and bound the exponential by $1 + \eta b_v(P,f(x_t))(y_t-P)$.
Fortunately, the following lemma shows that there exists a different $\widetilde b^\eta_v(p,q)$ function such that $1 +  \widetilde b^\eta_v(P,f(x_t))(y_t-P)$ upper bounds the exponential.

\begin{lemma}\label{lem:bernoulli-factor}
For any $p\in[0,1]$, $y\in\{0,1\}$, $h\in[-1,1]$, and $0<\eta\le1/2$,
define $F^\eta_{p,y}(h)=\exp\rbr{\eta h(y-p)-\eta^2p(1-p)h^2}$ and $S_p^\eta(h)=F^\eta_{p,1}(h)-F^\eta_{p,0}(h)$.
We have
$
F^\eta_{p,y}(h) \leq 1 + S_p^\eta(h)(y-p).
$
\end{lemma}
\begin{proof}
Since $F^\eta_{p,y}(h) = \E_{Y\sim\Ber(p)}\sbr{F^\eta_{p,Y}(h)} + S_p^\eta(h)(y-p)$, it suffices to show $\E_{Y\sim\Ber(p)}\sbr{F^\eta_{p,Y}(h)} \leq 1$.
Indeed, this holds by using the fact $e^x\le1+x+x^2$ for $|x|\le1/2$, with $X=\eta h(Y-p)$:
\begin{align*}
\E_{Y\sim\Ber(p)}\sbr{\exp\rbr{\eta h(Y-p)}}
&= \E_{Y\sim\Ber(p)}\sbr{e^X} 
\leq 1 + \E_{Y\sim\Ber(p)}\sbr{X} + \E_{Y\sim\Ber(p)}\sbr{X^2} \\
&= 1 + \eta^2 p(1-p)h^2 \leq \exp\rbr{\eta^2 p(1-p)h^2}. \qedhere
\end{align*}
\end{proof}
Therefore, if we define
\begin{equation}
  \widetilde b^\eta_v(p,q)=S_p^\eta(b_v(p,q)) = F^\eta_{p,1}(b_v(p,q))-F^\eta_{p,0}(b_v(p,q)),
  \label{eq:corrected-threshold-test}
\end{equation}
then continuing from \pref{eq:W_ratio_Jensen2} and using \pref{lem:bernoulli-factor}, we obtain $\frac{W_{t+1,f}(v)}{W_{t,f}(v)} \leq \E_{P\sim D_t}\sbr{F_{P,y_t}^\eta(b_v(P, f(x_t)))} \leq 1 + \E_{P\sim D_t}\sbr{\widetilde b^\eta_v(P,f(x_t))(y_t-P)}$, replacing $\eta b_v(P,f(x_t))$ in the derivation of the proof of \pref{lem:lipschitz-comparison}
with $\widetilde b^\eta_v(P,f(x_t))$.
This means that if we accordingly define the expected test function as
\begin{equation}
  G_t(p)=\frac{1}{\Phi_t}\E_{(f,v)\sim\pi_\delta}
  \sbr{W_{t,f}(v)\widetilde b^\eta_v(p,f(x_t))},
  \label{eq:mixable-test}
\end{equation}
then the whole analysis goes through in the same way.
This explains the design of our algorithm, shown in \pref{alg:mixable}.

\begin{algorithm}[t]
  \caption{An online omnipredictor for all mixable proper losses}
  \label{alg:mixable}
  \textbf{Input}: a horizon $T$, a finite prediction class $\calF$, a learning rate $0<\eta\le1/2$, and an EVI solver.  

  \textbf{Initialization:} $\delta=1/\rbr{2(T+1)}$, $I_\delta=[\delta,1-\delta]$, $M_\delta=2\log(2T+1)$, prior $\pi_\delta=\operatorname{Unif}(\calF)\otimes\mu_\delta$ where $\mu_\delta(\dmeas v)=\frac{\dmeas v}{M_\delta v(1-v)}$, $\epsilon_t=1/T$, and $W_{1,f}(v)=1$ for all $f\in\calF$ and $v\in I_\delta$.

  \For{$t=1,\ldots,T$}{
    Nature reveals $x_t$.

    Compute total mass $\Phi_t=\E_{(f,v)\sim\pi_\delta}[W_{t,f}(v)]$ and the expected test function $G_t$ as in \pref{eq:mixable-test}.\;

    Use the EVI solver to find a distribution $D_t$ over $I_\delta$, satisfying
    \[
      \E_{P\sim D_t}[G_t(P)(y-P)]\le\epsilon_t,
      \qquad\text{for all } y\in \{0,1\}.
    \]

    Nature reveals outcome $y_t$.

    For every expert $(f,v)\in\calF\times I_\delta$, define its reward as $z_{t,f,v}=\E_{P\sim D_t}[b_v(P,f(x_t))(y_t-P)]$.

    Update the weight via an exponential update with a (new) second-order correction term:
    \[
      W_{t+1,f}(v)=W_{t,f}(v)\exp\rbr{\eta z_{t,f,v}-\eta^2\E_{P\sim D_t}[P(1-P)b_v(P,f(x_t))^2]}.
    \]
  }
\end{algorithm}

\paragraph{Regret analysis.}
The discussion above directly implies the following second-order regret bound which is analogous to \pref{lem:lipschitz-comparison} but, importantly, contains the $P(1-P)$ factor (proofs for all lemmas can be found in \pref{app:factors}).
\begin{lemma}
  \label{lem:mixable-comparison}
  If $\ell$ is proper and $\beta$-mixable, then for any 
  $f\in\calF$, \pref{alg:mixable} satisfies
  \[
    \begin{aligned}
      \sum_{t=1}^T\int_{I_\delta}w_\ell(v)z_{t,f,v}\dmeas v
      \le{}&\order\rbr{\frac{\log(T)\log(|\calF|)}{\beta\eta}}
      +\eta\sum_{t=1}^T\E_{P\sim D_t}\int_{I_\delta}w_\ell(v)P(1-P)b_v(P,f(x_t))^2\dmeas v.
    \end{aligned}
  \]
\end{lemma}

Using the extra $P(1-P)$ factor, we then prove the following lemma that is analogous to \pref{lem:lipschitz-cancellation} and shows how the second-order term (of order $\eta$) and the negative properness gain together lead to a term of order $\eta^2$, even when the curvature $w_\ell(v)$ scales as $\frac{1}{v(1-v)}$.
\begin{lemma}
  \label{lem:mixable-cancellation}
  For any $\beta$-mixable proper loss $\ell$, $\eta\in(0,1/2]$, 
$p\in I_\delta$, and $q\in[0,1]$, we have
  \[
    \int_{I_\delta}w_\ell(v)\sbr{\eta p(1-p)b_v(p,q)^2-d_v(p,q)}\dmeas v
    \le\frac{\eta^2}{\beta}.
  \]
\end{lemma}
On the other hand, recall that we only consider thresholds and make predictions in $I_\delta$, whereas regret compares against the original benchmark reports in $[0,1]$.  
To address this, we only need to show that clipping the benchmark report to the interval $I_\delta$ only introduces a small approximation error of order $\order(\delta)$.

\begin{lemma}
  \label{lem:target-clipping}
  For any $\beta$-mixable proper loss $\ell$, 
  $q\in[0,1]$, and $y\in\{0,1\}$,
  \[
    \ell(\ClipOp(q,I_\delta),y)-\ell(q,y)
    \le-\frac{\log(1-\delta)}{\beta}\le\frac{2\delta}{\beta},
  \]
  where we recall $\ClipOp(q,I_\delta) = \min\{1-\delta, \max\{\delta, q\}\}$.
  Consequently, for every $p\in I_\delta$,
  \begin{equation}
    \ell(p,y)-\ell(q,y)
    \le\int_{I_\delta}w_\ell(v)\sbr{b_v(p,q)(y-p)-d_v(p,q)}\dmeas v
    +\frac{2\delta}{\beta}.
    \label{eq:truncated-comparison}
  \end{equation}
\end{lemma}
Since our algorithm only makes predictions in $I_\delta$, \pref{eq:truncated-comparison} justifies why it suffices to only consider thresholds in $I_\delta$.
Finally, combining all three lemmas and picking the optimal $\eta$
proves our main result \pref{thm:mixable}, that is, for every mixable proper loss $\ell$, \pref{alg:mixable} ensures
\begin{equation}
  \Reg_{\ell,\calF}(T)=\order\rbr{\frac{T^{1/3}(\log (T)\log(|\calF|))^{2/3}+\log T\log(|\calF|)}{\beta_\ell^\star}}.
  \label{eq:explicit-mixable}
\end{equation}
The details appear in \pref{app:mixable}.
We recall that this bound is optimal, as shown in \pref{thm:lower-bound-mixable} and \pref{app:lower}.

\paragraph{Deterministic forecasts.}
We remark that while we follow the framework of~\cite{farina2026blackbox} and present both \pref{alg:lipschitz} and \pref{alg:mixable} using a generic EVI solver that finds a randomize forecast $D_t$,
the EVI that these two algorithms need to solve actually admits a deterministic solution $D_t=\delta_{p_t}$ where $p_t$ is a root of the expected test function $G_t$ (that is, $G_t(p_t)=0$ and hence $\E_{P\sim \delta_{p_t}}[G_t(P)(y-P)]\le\epsilon_t$ trivially holds).
Such a root exists because $G_t$ is continuous and is nonnegative at the left endpoint and nonpositive at the right endpoint of the forecast interval (details omitted).
However, this is not the case for the best-of-all-worlds algorithm we will present in the next section (where $G_t$ could be discontinuous), which is also why we choose to present a generic randomized EVI solution for all algorithms.


\subsection{A best-of-all-worlds omnipredictor}
\label{sec:unified}
There are two drawbacks of \pref{alg:mixable}: first, it does not provide any regret guarantee beyond the class of mixable proper losses; second, it sacrifices the standard constant in $T$ regret for specific mixable losses such as squared loss and log loss, in exchange for a slower rate that holds for all mixable losses simultaneously.
In this section, we show that both drawbacks can be addressed via a best-of-all-worlds algorithm, which, as shown in \pref{thm:unified}, simultaneously achieves the best regret bounds for a finite class $\calLbd$ of general $[0,1]$-valued losses, a finite class $\calLmix$ of mixable proper losses, and the infinite class of all mixable proper losses.
Throughout the discussion, we will use a superscript $c \in \mathcal C=\{\mathrm{bd},\mathrm{mix},\mathrm{MIX}\}$ to distinguish similar quantities that are associated with these three different classes.

The key idea is still the same: treat each benchmark-loss pair $(f, \ell)$ as an expert (where $\ell$ can be from the two finite classes $\calLbd$ and $\calLmix$ or the V-shaped loss $\lambda_v$ for $v \in [\delta,1-\delta]$), run exponential weight updates over these experts, use the resulting distribution to compute an expected test function $G_t$, and finally solve the EVI involving $G_t$ to obtain the final randomized forecast $D_t$. 
For ease of presentation, it would be convenient to write the test function $G_t$ as a weighted average of three test functions $\{G_t^c\}_{c\in \mathcal C}$, each corresponding to one of the three different classes.

\paragraph{Component for $\calLbd$.}
This component is very similar to the original algorithm of~\citep{farina2026blackbox} where, based on the discussion in \pref{sec:algorithm_overview} and in particular \pref{eq:regret_identity_dropping_d}, we define the reward for each expert $(f,\ell)\in\calF\times\calLbd$ as $z^{\mathrm{bd}}_{t,f,\ell}=\E_{P\sim D_t}\sbr{b_\ell(P,f(x_t))(y_t-P)}$.
We then maintain a weight $W^{\mathrm{bd}}_{t,f,\ell}$ for each expert $(f, \ell)$ in round $t$.
For consistency, we update these weights in the same way as our \pref{alg:lipschitz} using the standard second-order correction with a learning rate $\eta^{\mathrm{bd}} \in (0, 1/4]$: 
\[
  W^{\mathrm{bd}}_{t+1,f,\ell}=W^{\mathrm{bd}}_{t,f,\ell}
  \exp\rbr{\eta^{\mathrm{bd}}z^{\mathrm{bd}}_{t,f,\ell}
  -(\eta^{\mathrm{bd}})^2(z^{\mathrm{bd}}_{t,f,\ell})^2}.
\]
The test function for this component also follows \pref{alg:lipschitz}: $G_t^{\mathrm{bd}}(p) = \frac{\eta^{\mathrm{bd}}}{\Phi_t^{\mathrm{bd}}}
  \E_{(f,\ell)\sim\pi^{\mathrm{bd}}}\sbr{W^{\mathrm{bd}}_{t,f, \ell} b_\ell(p,f(x_t))}$ where $\pi^{\mathrm{bd}}=\operatorname{Unif}(\calF) \otimes \operatorname{Unif}(\calLbd)$ is the uniform prior
  and $\Phi_t^{\mathrm{bd}} = \E_{(f,\ell)\sim \pi^{\mathrm{bd}}}\sbr{W^{\mathrm{bd}}_{t, f, \ell}}$ is the total mass for this component.

\paragraph{Component for all mixable losses.}
This component directly follows \pref{alg:mixable} where we define for each expert $(f,v) \in \calF \times I_\delta$,
a reward $z^{\mathrm{MIX}}_{t,f,v}=\E_{P\sim D_t}\sbr{b_v(P,f(x_t))(y_t-P)}$, a weight function updated via 
\[
  W^{\mathrm{MIX}}_{t+1,f}(v)=W^{\mathrm{MIX}}_{t,f}(v)
  \exp\rbr{\eta^{\mathrm{MIX}}z^{\mathrm{MIX}}_{t,f,v}
  -(\eta^{\mathrm{MIX}})^2\E_{P\sim D_t}\sbr{P(1-P)b_v(P,f(x_t))^2}}
\]
for a learning rate $\eta^{\mathrm{MIX}} \in (0, 1/2]$,
a prior $\pi^{\mathrm{MIX}}=\pi_\delta$ (\pref{eq:prior_delta}),
and a test function 
\begin{equation*}
  G^{\mathrm{MIX}}_t(p)=\frac{1}{\Phi_t^{\mathrm{MIX}}}\E_{(f,v)\sim\pi^{\mathrm{MIX}}}
  \sbr{W^{\mathrm{MIX}}_{t,f}(v)\widetilde b^{\eta^{\mathrm{MIX}}}_v(p,f(x_t))},
\end{equation*}
for total mass $\Phi_t^{\mathrm{MIX}} = \E_{(f,v)\sim \pi^{\mathrm{MIX}}}\sbr{W^{\mathrm{MIX}}_{t, f}(v)}$
and threshold test $\widetilde b^\eta_v$ as defined in \pref{eq:corrected-threshold-test}.

\paragraph{Component for $\calLmix$.}
For the last component that handles a fixed finite class $\calLmix$ of mixable proper losses,
instead of following the regret upper bound of \pref{eq:regret_identity_dropping_d},
we directly define the reward for each expert $(f,\ell)\in\calF\times\calLmix$ as the regret
\[
  z^{\mathrm{mix}}_{t,f,\ell}
  =\E_{P\sim D_t}\sbr{\ell(P,y_t)-\ell(\ClipOp(f(x_t),I_\delta),y_t)},
\]
where the clipping of the benchmark prediction avoids unboundedness issues (and is harmless due to \pref{lem:target-clipping}).
The weight functions are also updated in a simple manner via vanilla exponential weight updates with no correction terms (recall that $\beta_\ell^\star$ is the largest mixability constant for $\ell$):
\[
  W^{\mathrm{mix}}_{t+1,f,\ell}=W^{\mathrm{mix}}_{t,f,\ell}
  \exp\rbr{\beta_\ell^\star z^{\mathrm{mix}}_{t,f,\ell}}.
\]
To see what the test function should be, we introduce the following lemma that is analogous to \pref{lem:bernoulli-factor}.

\begin{lemma}
  \label{lem:mixability-factor}
  For any mixable proper loss $\ell$, $y\in \{0,1\}$, $p\in(0,1)$ and $q\in[0,1]$, define $J^y_\ell(p,q) = \exp\rbr{\beta_\ell^\star(\ell(p,y)-\ell(q,y))}$
  and $\overline{S}_\ell(p,q) = J^1_\ell(p,q)-J^0_\ell(p,q)$.
  We have $J^y_\ell(p,q) \leq 1+\overline{S}_\ell(p,q)(y-p)$.
\end{lemma}
\begin{proof}  
Since $J^y_\ell(p,q) = \E_{Y \sim \Ber(p) } \sbr{J^Y_\ell(p,q)}+\overline{S}_\ell(p,q)(y-p)$, it suffices to show 
  \begin{equation}
\E_{Y \sim \Ber(p) } \sbr{\exp\rbr{\beta_\ell^\star(\ell(p,Y)-\ell(q,Y))}} \leq 1,
    \label{eq:mixability-factor}
  \end{equation}
which is exactly the central condition of~\cite{van2015fast} and is implied by mixability (see their Proposition 4.12 and Theorem 3.10).
\end{proof}
With this lemma in hand, we can proceed similarly as before and bound the weight ratio as
\begin{align}
\frac{W^{\mathrm{mix}}_{t+1,f,\ell}}{W^{\mathrm{mix}}_{t,f,\ell}} 
&= \exp\rbr{\beta_\ell^\star z^{\mathrm{mix}}_{t,f,\ell}} 
\leq  \E_{P\sim D_t}\sbr{J^{y_t}_\ell(P, \ClipOp(f(x_t),I_\delta))} \tag{Jensen's inequality}\\
&\leq 1+ \E_{P\sim D_t}\sbr{\overline{S}_\ell(P,\ClipOp(f(x_t),I_\delta))(y_t-P)}. \tag{\pref{lem:mixability-factor}}
\end{align}
Therefore, following similar logic and defining $\overline{b}_\ell(p,q) = \overline{S}_\ell(p,\ClipOp(q,I_\delta))$,
we see that the test function in this case should be defined as
\begin{equation*}
  G^{\mathrm{mix}}_t(p)=\frac{1}{\Phi_t^{\mathrm{mix}}}\E_{(f,\ell)\sim\pi^{\mathrm{mix}}}
  \sbr{W^{\mathrm{mix}}_{t,f, \ell}\overline b_\ell(p,f(x_t))},
\end{equation*}
for a prior $\pi^{\mathrm{mix}}=\operatorname{Unif}(\calF) \otimes \operatorname{Unif}(\calLmix)$
and total mass $\Phi_t^{\mathrm{mix}} = \E_{(f,\ell)\sim \pi^{\mathrm{mix}}}\sbr{W^{\mathrm{mix}}_{t, f, \ell}}$.

\begin{algorithm}[!t]
  \caption{A best-of-all-worlds online omnipredictor}
  \label{alg:unified}
  \textbf{Input}: a horizon $T$, finite classes $\calF,\calLbd,\calLmix$, 
  learning rates $0<\eta^{\mathrm{bd}}\le1/4$ and $0<\eta^{\mathrm{MIX}}\le1/2$, and an EVI solver.  

  \textbf{Initialization:} Initial weight $W_1^c$ is $1$ for all experts and $c\in \{\mathrm{bd},\mathrm{mix},\mathrm{MIX}\}$, $\delta=1/\rbr{2(T+1)}$, $I_\delta=[\delta,1-\delta]$, and $\epsilon_t=1/T$.

  \For{$t=1,\ldots,T$}{
    Nature reveals $x_t$.

    Compute the expected test function $G_t$ as given by \pref{eq:master-factor}.\;

    Use the EVI solver to find a distribution $D_t$ over $I_\delta$, 
    satisfying
    \[
      \E_{P\sim D_t}\sbr{G_t(P)(y-P)}\le\epsilon_t,
      \qquad\forall  y\in \{0,1\}.
    \]
    Nature reveals outcome $y_t$.\;

    For every expert $(f,\ell)\in\calF\times\calLbd$, define its reward as
    \[
      z^{\mathrm{bd}}_{t,f,\ell}
      =\E_{P\sim D_t}\sbr{b_\ell(P,f(x_t))(y_t-P)}.
    \]
    Update the weights for the bounded loss component:
    \[
      W^{\mathrm{bd}}_{t+1,f,\ell}=W^{\mathrm{bd}}_{t,f,\ell}
      \exp\rbr{\eta^{\mathrm{bd}}z^{\mathrm{bd}}_{t,f,\ell}
      -(\eta^{\mathrm{bd}})^2(z^{\mathrm{bd}}_{t,f,\ell})^2}.
    \]
    For every expert $(f,\ell)\in\calF\times\calLmix$, 
    define its reward as
    \[
      z^{\mathrm{mix}}_{t,f,\ell}
      =\E_{P\sim D_t}\sbr{\ell(P,y_t)-\ell(\ClipOp(f(x_t),I_\delta),y_t)}.
    \]
    Update the weights for the finite mixable loss component:
    \[
      W^{\mathrm{mix}}_{t+1,f,\ell}=W^{\mathrm{mix}}_{t,f,\ell}
      \exp\rbr{\beta_\ell^\star z^{\mathrm{mix}}_{t,f,\ell}}.
    \]
    For every expert $(f,v)\in\calF\times I_\delta$, define its reward as
    \[
      z^{\mathrm{MIX}}_{t,f,v}
      =\E_{P\sim D_t}\sbr{b_v(P,f(x_t))(y_t-P)}.
    \]
    Update the weights for the universal mixable loss component:
    \[
      W^{\mathrm{MIX}}_{t+1,f}(v)=W^{\mathrm{MIX}}_{t,f}(v)
      \exp\rbr{\eta^{\mathrm{MIX}}z^{\mathrm{MIX}}_{t,f,v}
      -(\eta^{\mathrm{MIX}})^2\E_{P\sim D_t}\sbr{P(1-P)b_v(P,f(x_t))^2}}.
    \]
  }
\end{algorithm}

\paragraph{Combining everything.}
Finally, we combine all three test functions by taking a weighted average
\begin{equation}
  G_t(p)=\frac{1}{\Phi_t}\sum_{c\in\mathcal C}
  \Phi_t^c G^c_{t}(p),
  \label{eq:master-factor}
\end{equation}
where $\Phi_t = \sum_{c\in\mathcal C} \Phi_t^c$,
and solve the corresponding EVI, that is, find a distribution $D_t$ over $I_\delta$ such that $\E_{P\sim D_t}\sbr{G_t(P)(y-P)}\le\epsilon_t$ for any $y\in\{0,1\}$ and $\epsilon_t = 1/T$.
The complete algorithm is shown in \pref{alg:unified}.

\paragraph{Regret analysis.}
The analysis mostly follows previous discussion and shows the key step $\Phi_{t+1} \leq (1+\epsilon_t)\Phi_t$.
Eventually, picking the learning rates appropriately proves
that \pref{alg:unified} achieves 
   $
        \Reg_{\ell,\calF}(T)
        =
        \order\rbr{\frac{\log\rbr{|\calF||\calLmix|}}{\beta_{\ell}^\star}}
    $ for any $\ell \in \calLmix$,
    $
        \Reg_{\ell,\calF}(T)
        =\otil\rbr{\frac{T^{1/3}(\log|\calF|)^{2/3}}{\beta_\ell^\star}}
    $
    for any mixable proper loss $\ell$ (whether or not it belongs to $\calLmix$),
    and $
        \Reg_{\ell,\calF}(T)
        =
        \order
        \rbr{
          \sqrt{T\log\rbr{|\calF||\calLbd|}}
        }$
    for any $\ell\in\calLbd$. See \pref{app:unified} for the full proof.
    Note that since Lipschitz proper losses are mixable (\pref{lem:automatic-curvature}), \pref{alg:unified} also achieves the same result for Lipschitz proper losses as \pref{alg:lipschitz} does (up to $\log T$ factors).

    In \pref{cor:unified}, we instantiate $\calLbd$ and $\calLmix$ with concrete losses and obtain regret bounds for more specific loss families.
    For example, we instantiate $\calLmix$ with the squared loss and the log loss, two of the most commonly used proper losses.
    We also instantiate $\calLbd$ with some threshold loss functions that serve as approximate bases for both Lipschitz losses (that are not necessarily proper) and generalized linear model (GLM) losses, so that our regret bound against any loss in $\calLbd$ directly implies a regret bound for losses in those two families too.
    In particular, the approximate bases we use for the GLM class appear to be unknown in the literature and could be of independent interest; see \pref{prop:glm-threshold-approximation}.

\section{Conclusion}
\label{sec:conclusion}

We studied whether online omniprediction guarantees for finite loss classes extend to infinite classes.
We showed that sublinear regret is impossible for all bounded proper losses, even with only two benchmarks.
In contrast, for all mixable proper losses, we achieve $\otil(T^{1/3}(\log|\calF|)^{2/3}/\beta_\ell^\star)$ regret simultaneously, with the $T^{1/3}$ dependence being optimal up to logarithmic factors even for Lipschitz proper losses.
This reveals a separation between the full class of mixable proper losses and finite collections, for which constant-in-$T$
regret is achievable.

Our approach combines the V-shaped representation of proper losses with continuous exponential weights and second-order corrections that exploit the negative properness gain.
A Bernoulli-variance correction further handles the potentially unbounded curvature of mixable losses.
By combining this approach with finite-loss components, we obtain a single forecaster that retains the existing regret guarantees for finite classes of mixable and bounded losses, while simultaneously achieving our guarantee for all mixable proper losses.
Suitable finite bases also extend these guarantees to
all Lipschitz and GLM losses.

\section*{Disclosure of Generative AI Use}

We used GPT-6 as an auxiliary tool in the
development of this paper. In particular, it was used
to assist with exploring algorithmic ideas, developing
and refining proof arguments, identifying connections
to related literature, and improving the clarity and
presentation of the manuscript.
The authors independently verified the correctness,
originality, and presentation of all results, proofs,
and references reported in the paper, and take full
responsibility for the final content.

\newpage
\appendix

\section{Proofs of Main Results}
\label{app:main}


\subsection{Impossibility for all bounded proper losses}
\label{app:impossibility}

\impossibility*

\begin{proof}[{Proof of \protect\pref{thm:impossibility}}]
The proof builds on the interval-bisection construction used by
\cite[Theorem~8.2]{rakhlin2014statistical} to show that threshold
classifiers are not online learnable. Their construction generates
independent fair outcomes and chooses each context from a binary tree
according to the previous outcomes. Every resulting path can be
perfectly classified by a threshold chosen in hindsight, even though
the next outcome cannot be predicted. We use this same construction.
The additional work is to obtain a lower bound with only two fixed
benchmarks. We encode the threshold in the proper loss and use a second
copy of the tree, shifted by $1/4$. Each copy gives a loss under which
its corresponding benchmark has zero loss. The separation between the
copies then forces the learner to incur linear expected loss for at
least one of them.

Fix an arbitrary integer $T\ge1$. The context space is $\calX=[0,3/4]$,
and the fixed benchmark class is $\calF=\{f_1,f_2\}$, where
$f_1(x)=x$ and $f_2(x)=x+1/4$.
We use a binary interval tree of depth $T-1$ to specify the loss family.
Starting from $J_\varnothing=(1/3,5/12)$, for each binary string
$h\in\{0,1\}^{\le T-1}$, let $J_h$ be the interval obtained
by successively selecting the lower half for each bit $1$
and the upper half for each bit $0$.
Let $m_h$ be the midpoint of $J_h$.
For each leaf $r\in\{0,1\}^{T-1}$, let $v_{r,0}$ and $v_{r,1}$
be the midpoints of the upper and lower halves of $J_r$,
respectively. Thus,
$  v_{r,1}<m_r<v_{r,0}.$
Using two thresholds inside each leaf allows the same scalar
context $m_r$ to support either outcome on the final round.
Using the V-shaped proper losses defined in \pref{sec:prelim}
we define the family
\[
  \calL_T
  =\{\ell_{r,b}:r\in\{0,1\}^{T-1},\ b\in\{0,1\}\},
  \qquad
  \ell_{r,b}=\lambda_{v_{r,b}+b/4}.
\]
The leaf intervals are disjoint, and the two sets of thresholds lie
in $(1/3,5/12)$ and $(7/12,2/3)$. Thus $\calL_T$ contains exactly
$2^T$ distinct $[0,1]$-valued proper losses.

The hard distribution chooses $r\sim\Unif(\{0,1\}^{T-1})$ and an
independent $b\sim\Ber(1/2)$. Writing
$r_{<t}=(r_1,\ldots,r_{t-1})$, the selected sequence and evaluation
loss are
\[
  (x_t,y_t)=
  \begin{cases}
    (m_{r_{<t}},r_t),&1\le t<T,\\
    (m_r,b),&t=T,
  \end{cases}
  \qquad
  \ell=\ell_{r,b}.
\]

Here, the final outcome $b$ also selects the evaluation loss.
Each pair $(r,b)$ therefore specifies a deterministic sequence and
one evaluation loss.
When $T=1$, $r$ is the empty string and only the final round remains.
Fix an arbitrary, possibly randomized forecaster $A$, whose randomness
is independent of the hard distribution. Recall that $D_t$ is its forecast
distribution given the history and current context. For each fixed
pair $(r,b)$, we define
\[
  R(A;r,b)
  =\E_A\sbr{\Reg_{\ell_{r,b},\calF}(T)},
\]
where $\E_A$ averages over all randomness of the forecaster.
Since the expectation over $P\sim D_t$ is already part of the regret
definition, $R(A;r,b)$ is a deterministic number for each fixed
instance.

Since both $v_{r,0}$ and $v_{r,1}$ lie in $J_r$, interval nesting gives
$\Ind{m_{r_{<t}}\ge v_{r,b}}=r_t$ for every $t<T$, with no ties.
Thus $f_{b+1}$ has zero loss on the first $T-1$ rounds.
It also has zero loss on the final round. If $b=0$, then
$f_1(x_T)=m_r<v_{r,0}$ and $y_T=0$. If $b=1$, then
$f_2(x_T)=m_r+1/4>v_{r,1}+1/4$ and $y_T=1$.
For these proper losses, we use the identity post-processing
$\pi_{\lambda_v}(p)=p$, and have that 
\[
  R(A;r,b)
  =\E_A\sbr{
    \sum_{t=1}^T\E_{P\sim D_t}
      \lambda_{v_{r,b}+b/4}(P,y_t)
  }.
\]

For all $v_0,v_1\in(1/3,5/12)$, $p\in[0,1]$, and $y\in\{0,1\}$,
\begin{equation}
  \lambda_{v_0}(p,y)+\lambda_{v_1+1/4}(p,y)
  \ge\frac13\sbr{
    (1-y)\Ind{p\ge1/2}+y\Ind{p<1/2}
  }.
  \label{eq:impossibility-pair}
\end{equation}
Indeed, when $p\ge1/2$, the lower-threshold loss is at least $1/3$
for $y=0$. When $p<1/2$, the upper-threshold loss is at least $1/3$
for $y=1$. These comparisons also hold for any fixed Bayes-optimal
tie rule, since the contributing prediction lies strictly on one
side of the corresponding threshold.

For every $t<T$, $D_t$ depends only on $r_{<t}$ and the forecaster's
randomness, and in particular does not depend on $b$.
The two choices of $b$ therefore give the same forecast distribution
when the forecaster uses the same random seed.
Conditional on the information available when $D_t$ is chosen,
$r_t$ remains an independent fair bit.
The right-hand side of \pref{eq:impossibility-pair} does not depend
on either leaf threshold. Averaging first over $b$ and then over
the outcomes and the forecaster's randomness gives
\begin{align*}
  &\E_{r,b}\E_A\E_{P\sim D_t}
    \lambda_{v_{r,b}+b/4}(P,r_t)\\
  &\qquad=
    \frac12\E_r\E_A\E_{P\sim D_t}
    \sbr{\lambda_{v_{r,0}}(P,r_t)+\lambda_{v_{r,1}+1/4}(P,r_t)}\\
  &\qquad\ge
    \frac16\E_r\E_A\E_{P\sim D_t}
    \sbr{(1-r_t)\Ind{P\ge1/2}+r_t\Ind{P<1/2}}
    =\frac1{12}.
\end{align*}

On the final round, $D_T$ may depend on $r$ and the forecaster's
randomness, but not on $b$. For every $p\in[0,1]$, the same
separation of thresholds gives
\[
  \frac12\sbr{
    \lambda_{v_{r,0}}(p,0)+\lambda_{v_{r,1}+1/4}(p,1)
  }\ge\frac16.
\]
Averaging over $b$, the forecaster's randomness, and $P\sim D_T$
therefore gives expected loss at least $1/6$ on this round.
Consequently, the arbitrary randomized forecaster $A$ satisfies
\[
  \E_{r,b}R(A;r,b)
  \ge\frac{T-1}{12}+\frac16
  \ge\frac{T}{12}.
\]

Since this is an average over finitely many pairs $(r,b)$, there
exists a fixed pair $(r^*,b^*)$ such that
$R(A;r^*,b^*)\ge T/12$.
For the deterministic sequence specified by this pair,
\[
  \sup_{\ell\in\calL_T}
    \E_A\sbr{\Reg_{\ell,\calF}(T)}
  \ge
    \E_A\sbr{\Reg_{\ell_{r^*,b^*},\calF}(T)}
  =R(A;r^*,b^*)
  \ge\frac{T}{12}.
\]
The pair is selected after taking the expectation over the
forecaster's randomness. Thus both the sequence and the single loss
$\ell_{r^*,b^*}$ are fixed independently of its realized random seed,
as required.
\end{proof}

\subsection{Lipschitz and mixable lower bounds for every horizon}
\label{app:lower}

\mixableLB*

\begin{proof}[{Proof of \protect\pref{thm:lower-bound-mixable}}]
We first establish an $\Omega(LT^{1/3})$ lower bound for bounded
$L$-Lipschitz proper losses. Throughout the proof, $L>0$,
the integer $T\ge1$, and the forecaster is arbitrary and fixed.
The context space is $\calX=[0,1]^2$, and the fixed benchmark
class is $\calF=\{f_+,f_-\}$, where
$f_+(x)=x_1$ and $f_-(x)=x_2$.

\textbf{A construction with $K$ phases.}
We partition total $T$ rounds into $K$ phases, with round-index sets
$\mathcal I_j$ and lengths
$n_j\in\{\lfloor T/K\rfloor,\lceil T/K\rceil\}$.
Let $v_j=0.45+\frac{j-1/2}{10K},
h_j=\frac{1}{100\sqrt{n_j}}$.
In phase $j \in [K]$, the environment reveals
$x_t=(v_j+h_j,v_j-h_j)$ and draws
$y_t\sim\Ber(v_j)$, independently across all rounds.
Thus the two benchmarks predict $v_j+h_j$ and $v_j-h_j$.
We will choose a value of $K$ such that the intervals
$[v_j-h_j,v_j+h_j]$ are pairwise disjoint.
Let $S_j=\sum_{t\in\mathcal I_j}(y_t-v_j)$ be the fluctuation of outcomes in phase $j$.
For shorthand, we denote $\sigma_j^2=n_jv_j(1-v_j).$
Since $v_j\in[0.45,0.55]$, the Bernoulli moment formulas (e.g.~\citep[Table~2]{skorski2020handy}) give
\[
\E S_j^2=\sigma_j^2,
\qquad
\E S_j^4
=3\sigma_j^4+\sigma_j^2(1-6v_j(1-v_j))
\le3\sigma_j^4.
\]
H\"older's inequality therefore yields 
\[
\mathbb{E}[S_j^2]
= \mathbb{E}\bigl[|S_j|^{2/3}|S_j|^{4/3}\bigr]
\leq \bigl(\mathbb{E}|S_j|\bigr)^{2/3}
     \bigl(\mathbb{E}[S_j^4]\bigr)^{1/3},
\]
and thus
\begin{equation} \label{eq:LB_ESj}
  \E|S_j|
\ge
\frac{(\E S_j^2)^{3/2}}{(\E S_j^4)^{1/2}}
\ge \frac{\sigma_j}{\sqrt{3}} \ge \frac{\sqrt{n_j}}4.
\end{equation}

\textbf{A fixed family of Lipschitz proper losses.}
For each phase $j\in[K]$ and subset $U\subseteq[K]$, we define
\[
w_j(v)=L\Ind{|v-v_j|\le h_j},
\qquad
w^U(v)=\sum_{j\in U}w_j(v).
\]
Disjointness of the intervals gives $0\le w^U\le L$ for any $U\subseteq[K]$.
Write $\ell_w$ for the proper loss associated with density $w$
through \pref{eq:v-shaped-rep}, with $C_\ell(y)=0$.
Since $w^U(v)\le L\le L/\max\{v,1-v\}$,
\pref{lem:automatic-curvature} gives
$\ell_{w^U}\in\LipLosses{L}$.
Moreover,
\[
0\le\ell_{w^U}(q,y)
\le L\int_0^1\lambda_v(q,y)\dmeas v
\le L/2.
\]
The family $\{\ell_{w^U}:U\subseteq[K]\}$ is fixed
independently of the outcomes.

\textbf{Two complementary losses.}
For a realized outcome sequence, let
\[
U_+=\{j:S_j\ge0\},
\qquad
U_-=[K]\setminus U_+,
\qquad
\ell_\pm=\ell_{w^{U_\pm}},
\]
and
\[
R_s=
\sum_{t=1}^T
\E_{P\sim D_t}
\bigl[
\ell_s(P,y_t)-\ell_s(f_s(x_t),y_t)
\bigr],
\qquad \text{where}\qquad  s\in\{+,-\}.
\]
Thus $R_s$ is the cumulative loss difference relative to the
fixed benchmark $f_s$, and
$\Reg_{\ell_s,\calF}(T)\ge R_s$.
Although $\ell_+$ and $\ell_-$ depend on the outcomes, their sum
is the fixed proper loss
\[
\bar\ell=\ell_++\ell_-=\ell_{\bar w},
\qquad
\bar w(v)=\sum_{j=1}^K w_j(v).
\]
For each round $t\in\mathcal I_j$, conditional on the history and
the choice of $D_t$, the outcome $y_t$ still follows
$\Ber(v_j)$. Properness of $\bar\ell$ gives
\[
\E_{y_t\sim\Ber(v_j)}
\E_{P\sim D_t}
\bigl[
\bar\ell(P,y_t)-\bar\ell(v_j,y_t)
\bigr]
\ge0.
\]
Adding and subtracting the losses at $v_j$ therefore yields
\begin{align*}
\E[R_++R_-]
\ge
\sum_{j=1}^K\sum_{t\in\mathcal I_j}
\sum_{s\in\{+,-\}}
\E \bigl[
\ell_s(v_j,y_t)-\ell_s(f_s(x_t),y_t)
\bigr],
\end{align*}
where the expectation includes the outcomes and any forecaster
randomness.
For every $t\in\mathcal I_j$, subtracting the integral representations
of $\ell_{w_j}$ at $v_j$ and $v_j\pm h_j$ gives
\begin{align*}
&\ell_{w_j}(v_j,y_t)-\ell_{w_j}(v_j\pm h_j,y_t)\\
&\qquad=\int_{v_j}^{v_j\pm h_j}(y_t-v)w_j(v)\dmeas v\\
&\qquad=L\int_{v_j}^{v_j\pm h_j}
  \sbr{(y_t-v_j)-(v-v_j)}\dmeas v\\
&\qquad=L\rbr{\pm h_j(y_t-v_j)-\frac{h_j^2}{2}}.
\end{align*}
The second equality uses $w_j(v)=L$ throughout
$[v_j-h_j,v_j+h_j]$.
Summing over $t\in\mathcal I_j$ and using
$S_j=\sum_{t\in\mathcal I_j}(y_t-v_j)$ and $|\mathcal I_j|=n_j$ yields
\begin{align*}
\sum_{t\in\mathcal I_j}
\rbr{\ell_{w_j}(v_j,y_t)-\ell_{w_j}(v_j\pm h_j,y_t)}
=
L\rbr{\pm h_jS_j-\frac{n_jh_j^2}{2}}.
\end{align*}
When $S_j\ge0$, this contribution appears in the comparison
between $\ell_+$ and $f_+$; otherwise, it appears in the
comparison between $\ell_-$ and $f_-$.
Every other density $w_k$, with $k\ne j$, vanishes between
$v_j$ and either benchmark prediction, so its contribution
to these loss differences is zero. Consequently,
\begin{align*}
\E[R_++R_-]
&\ge
L\sum_{j=1}^K
\rbr{h_j\E|S_j|-\frac{n_jh_j^2}{2}}\\
&\ge
LK\rbr{\frac1{400}-\frac1{20000}}
\ge\frac{LK}{500},
\end{align*}
where the second inequality uses \pref{eq:LB_ESj}.

\textbf{Choosing $K$ and a fixed witness.}
The preceding lower bound increases with $K$. Since
$\E|S_j|=\Theta(\sqrt{n_j})$, our choice
$h_j=1/(100\sqrt{n_j})$ ensures
$h_j\E|S_j|=\Theta(1)$ in each phase.
Moreover, $n_j\asymp T/K$ implies
$h_j\asymp\sqrt{K/T}$.
As $v_{j+1}-v_j=1/(10K)$, keeping the intervals
$[v_j-h_j,v_j+h_j]$ pairwise disjoint requires
$\sqrt{K/T}=O(1/K)$, or equivalently,
$K=O(T^{1/3})$.
We therefore choose $K=\lfloor T^{1/3}\rfloor$.
Since $T\ge K^3$, we have $n_j\ge K^2$ and
$h_j\le1/(100K)$, ensuring that these intervals
are pairwise disjoint and contained in $[0.45,0.55]$.

The selected losses $\ell_\pm$ depend only on the
outcomes, not on the forecaster's randomness.
Taking expectation over the forecaster's randomness
and then averaging over the outcomes, there exist
a deterministic outcome sequence and $s\in\{+,-\}$
such that
\[
\E[\Reg_{\ell_s,\calF}(T)]
\ge \E[R_s]
\ge \frac{LK}{1000}
\ge \frac{LT^{1/3}}{2000},
\]
where the expectations are now only over the
forecaster's randomness and the last inequality
uses $\lfloor T^{1/3}\rfloor\ge T^{1/3}/2$.
Thus, both the context--outcome sequence and
the loss $\ell_s$ are fixed independently of the
forecaster's randomness.

Finally,
$\LipLosses{2/\beta}\subseteq\MixLosses{\beta}$
by \pref{lem:automatic-curvature}.
Taking $L=2/\beta$ gives a bounded $2/\beta$-Lipschitz,
$\beta$-mixable proper loss with expected regret
$\Omega(T^{1/3}/\beta)$.
\end{proof}

\subsection{The universal mixable upper bound}
\label{app:mixable}

\mixable*

\begin{proof}[{Proof of \protect\pref{thm:mixable}}]
  We consider an arbitrary mixable proper loss $\ell$.
  If $\beta_\ell^\star=+\infty$, \pref{lem:automatic-curvature}
  gives $w_\ell=0$ almost everywhere, so $\ell$ is independent
  of the prediction and its regret is zero.
  Otherwise, we take $\beta=\beta_\ell^\star$.
  The mixability characterization in \pref{lem:automatic-curvature}
  then gives $\ell\in\MixLosses{\beta}$.
  The learning rate $0<\eta\le1/2$ and predictor $f\in\calF$
  are arbitrary and fixed.
  Averaging \pref{lem:target-clipping} and
  \pref{lem:mixable-cancellation} over $D_t$ and applying
  \pref{lem:mixable-comparison} gives
  \[
    \sum_{t=1}^T\E_{P\sim D_t}
    [\ell(P,y_t)-\ell(f(x_t),y_t)]
    \le\order\rbr{
      \frac1\beta\cbr{
        \frac{M_\delta\log(e|\calF|)}\eta
        +T\eta^2+T\delta
      }
    }.
  \]
  Balancing the complexity and variance terms gives the choice
  \begin{equation}
    \eta=\min\cbr{
      \frac12,
      \rbr{\frac{M_\delta\log(e|\calF|)}{T}}^{1/3}
    }.
    \label{eq:mixable-rate}
  \end{equation}
  Since $M_\delta=2\log(2T+1)$ and $T\delta\le1/2$,
  comparison with the best $f\in\calF$ yields
  \[
    \Reg_{\ell,\calF}(T)
    \le\order\rbr{\frac{
      T^{1/3}(\log(T)\log(|\calF|))^{2/3}
      +\log(T)\log(|\calF|)
    }{\beta}}.
  \]
  Since $\beta=\beta_\ell^\star$, this gives the claimed
  dependence on the mixability constant. 
\end{proof}

\subsection{The best-of-all-worlds guarantee}
\label{app:unified}

\unified*
\begin{proof}[{Proof of \protect\pref{thm:unified}}]
  The learning rates satisfy $0<\eta^{\mathrm{bd}}\le1/4$ and $0<\eta^{\mathrm{MIX}}\le1/2$.
  If a loss has $\beta_\ell^\star=+\infty$, \pref{lem:automatic-curvature} makes it independent of the prediction, so its regret is zero. For these losses, the factors $J_\ell^y$ are taken to be one, $\overline b_\ell=0$, and the weights $W^{\mathrm{mix}}_{t,f,\ell}$ remain one.

  We first bound the total weight $\Phi_t$ using the component weights and test functions defined in \pref{sec:unified}.
  For the bounded component, every $\ell\in\calLbd$ takes
values in $[0,1]$, so $|\Delta_\ell(q)|\le1$ and hence
$|b_\ell(p,q)|\le2$.
Since $|y_t-P|\le1$, this gives
$|z^{\mathrm{bd}}_{t,f,\ell}|\le2$.
  Thus $|\eta^{\mathrm{bd}}z^{\mathrm{bd}}_{t,f,\ell}|\le1/2$, and $e^{x-x^2}\le1+x$ for $|x|\le1/2$ gives
  \[
    \begin{aligned}
      \frac{W^{\mathrm{bd}}_{t+1,f,\ell}}{W^{\mathrm{bd}}_{t,f,\ell}}
      &\le1+\eta^{\mathrm{bd}}z^{\mathrm{bd}}_{t,f,\ell}=1+\eta^{\mathrm{bd}}\E_{P\sim D_t}
        [b_\ell(P,f(x_t))(y_t-P)].
    \end{aligned}
  \]
  Multiplying by $W^{\mathrm{bd}}_{t,f,\ell}$ and averaging over $\pi^{\mathrm{bd}}$ gives the bound below for $c=\mathrm{bd}$.
  For $c=\mathrm{MIX}$, the same bound follows from Jensen's inequality and \pref{lem:bernoulli-factor}, as in the proof of \pref{lem:mixable-comparison}.
  For $c=\mathrm{mix}$, it follows from Jensen's inequality and \pref{lem:mixability-factor}.
  In each case, the definition of $G_t^c$ therefore gives that for each $c\in\mathcal C$
  \[
    \Phi_{t+1}^c
    \le\Phi_t^c\rbr{1+\E_{P\sim D_t}[G_t^c(P)(y_t-P)]}
 .
  \]
  The same factor bounds give $1+G_t^c(p)(y-p)\ge0$ for $c=\mathrm{mix},\mathrm{MIX}$.
  For $c=\mathrm{bd}$, this follows from $|G_t^{\mathrm{bd}}(p)|\le2\eta^{\mathrm{bd}}\le1/2$.
  Their weighted average $G_t$ in \pref{eq:master-factor} consequently satisfies $1+G_t(p)(y-p)\ge0$.
  Since $\delta/(1-\delta)=1/(2T+1)<\epsilon_t=1/T$, \pref{lem:interior-evi} supplies the required forecast distribution.

  Summing the component inequalities and applying \pref{eq:master-factor} and the EVI condition yields
  \[
    \Phi_{t+1}
    \le\Phi_t\rbr{1+\E_{P\sim D_t}[G_t(P)(y_t-P)]}
    \le(1+\epsilon_t)\Phi_t.
  \]
  Each component starts with weights equal to one and has a probability prior, so $\Phi_1^c=1$ and $\Phi_1=3$.
  Therefore
  \[
    \Phi_{T+1}^c\le\Phi_{T+1}
    \le3\prod_{t=1}^T(1+\epsilon_t)\le3e
    \qquad(c\in\mathcal C).
  \]
  
  The three guarantees follow in the order stated in \pref{thm:unified}.
  For the finite mixable component, $(f,\ell)\in\calF\times\calLmix$ is arbitrary and fixed, with $0<\beta_\ell^\star<+\infty$.
  The prior $\pi^{\mathrm{mix}}$ assigns mass $1/(|\calF||\calLmix|)$ to this expert, so
  \[
    \frac{W^{\mathrm{mix}}_{T+1,f,\ell}}{|\calF||\calLmix|}
    \le\Phi_{T+1}^{\mathrm{mix}}\le3e.
  \]
  Taking logarithms and expanding the update in \pref{alg:unified} gives
  \[
    \sum_{t=1}^T z^{\mathrm{mix}}_{t,f,\ell}
    \le\frac{1+\log(3|\calF||\calLmix|)}{\beta_\ell^\star}.
  \]
  By the definition of $z^{\mathrm{mix}}_{t,f,\ell}$ and \pref{lem:target-clipping},
  \begin{align*}
    \sum_t\E_{P\sim D_t}
    \sbr{\ell(P,y_t)-\ell(f(x_t),y_t)}
 \le
    \sum_t z^{\mathrm{mix}}_{t,f,\ell}
    +\frac{2T\delta}{\beta_\ell^\star}\le
    \frac{1+\log(3|\calLmix|\,|\calF|)+2T\delta}{\beta_\ell^\star}.
  \end{align*}
  The use of $\beta_\ell^\star$ in the update and clipping bound is justified by \pref{lem:mixability-factor}, which includes the finite optimal constant. Since $T\delta\le1/2$, taking the best benchmark proves \pref{eq:unified-listed-mixable}.

  For the universal mixable component, $\ell$ is an arbitrary mixable proper loss and $f\in\calF$ is arbitrary and fixed.
  The case $\beta_\ell^\star=+\infty$ again gives zero regret.
  Otherwise, we take $\beta=\beta_\ell^\star$, which is allowed by the mixability characterization in \pref{lem:automatic-curvature}.
  The curvature mass is $m_\delta=\int_{I_\delta}w_\ell(v)\dmeas v\le M_\delta/\beta$.
  When $m_\delta>0$, we use the distribution $Q=\delta_f\otimes\nu_\delta$ on $\calF\times I_\delta$, where
  \[
    \nu_\delta(\dmeas v)=\frac{w_\ell(v)}{m_\delta}\dmeas v
    \quad(v\in I_\delta).
  \]
  As in the proof of \pref{lem:mixable-comparison}, the curvature bound and $\pi^{\mathrm{MIX}}=\pi_\delta$ yield
  \[
    \KL(Q\Vert\pi^{\mathrm{MIX}})
    =\log|\calF|+\KL(\nu_\delta\Vert\mu_\delta)
    \le\log|\calF|+\log\frac{M_\delta}{\beta m_\delta}.
  \]
  The argument of \pref{eq:prior-comparison}, applied to $W^{\mathrm{MIX}}_{T+1,f}(v)$ and its component prior, gives
  \[
    \begin{aligned}
      \int_{I_\delta}w_\ell(v)\log W^{\mathrm{MIX}}_{T+1,f}(v)\dmeas v
      &\le m_\delta\sbr{\KL(Q\Vert\pi^{\mathrm{MIX}})
        +\log\Phi_{T+1}^{\mathrm{MIX}}}\\
      &\le m_\delta\sbr{1+\log(3|\calF|)
        +\log\frac{M_\delta}{\beta m_\delta}}\\
      &\le\frac{M_\delta}{\beta}(1+\log(3|\calF|)+1/e).
    \end{aligned}
  \]
  The second inequality uses $\Phi_{T+1}^{\mathrm{MIX}}\le3e$.
  The last uses $m_\delta\le M_\delta/\beta$ and $u\log(1/u)\le1/e$ for $u=\beta m_\delta/M_\delta\in(0,1]$.
  Expanding the log update in \pref{alg:unified} gives
  \[
    \begin{aligned}
      \sum_t\int_{I_\delta}w_\ell(v)z^{\mathrm{MIX}}_{t,f,v}\dmeas v
      \le{}&\frac{M_\delta(1+\log(3|\calF|)+1/e)}{\beta\eta^{\mathrm{MIX}}}\\
      &+\eta^{\mathrm{MIX}}\sum_t\E_{P\sim D_t}
      \int_{I_\delta}w_\ell(v)P(1-P)b_v(P,f(x_t))^2\dmeas v.
    \end{aligned}
  \]
  When $m_\delta=0$, the threshold integrals vanish, so this bound still holds.  By the definition of $z^{\mathrm{MIX}}_{t,f,v}$, \pref{eq:truncated-comparison} and \pref{lem:mixable-cancellation} then yield
  \begin{align*}
    &\sum_t\E_{P\sim D_t}
    \sbr{\ell(P,y_t)-\ell(f(x_t),y_t)}\\
    &\le
    \sum_t\int_{I_\delta}w_\ell(v)z^{\mathrm{MIX}}_{t,f,v}\dmeas v
    -\sum_t\E_{P\sim D_t}\int_{I_\delta}
    w_\ell(v)d_v(P,f(x_t))\dmeas v
    +\frac{2T\delta}{\beta}\\
    &\le
    \frac{M_\delta(1+\log(3|\calF|)+1/e)}
    {\beta\eta^{\mathrm{MIX}}}\\
    &\quad+
    \sum_t\E_{P\sim D_t}\int_{I_\delta}w_\ell(v)
    \sbr{\eta^{\mathrm{MIX}}P(1-P)b_v(P,f(x_t))^2
    -d_v(P,f(x_t))}\dmeas v
    +\frac{2T\delta}{\beta}\\
    &\le
    \frac{M_\delta(1+\log(3|\calF|)+1/e)}
    {\beta\eta^{\mathrm{MIX}}}
    +\frac{T(\eta^{\mathrm{MIX}})^2+2T\delta}{\beta}.
  \end{align*}
  With $\eta^{\mathrm{MIX}}$ chosen as in \pref{eq:mixable-rate}, comparison with the best $f\in\calF$ gives
  \[
    \Reg_{\ell,\calF}(T)
    \le\order\rbr{\frac{
      T^{1/3}(M_\delta\log(|\calF|))^{2/3}
      +M_\delta\log(|\calF|)}{\beta_\ell^\star}},
  \]
which proves \pref{eq:unified-universal-mixable}.
  For the bounded component, $(f,\ell)\in\calF\times\calLbd$ is arbitrary and fixed.
  Its prior mass under $\pi^{\mathrm{bd}}$ is $1/(|\calF||\calLbd|)$, so
  \[
    \frac{W^{\mathrm{bd}}_{T+1,f,\ell}}{|\calF||\calLbd|}
    \le\Phi_{T+1}^{\mathrm{bd}}\le3e.
  \]
  Taking logarithms and expanding the update gives
  \[
    \sum_t z^{\mathrm{bd}}_{t,f,\ell}
    \le\frac{1+\log(3|\calF||\calLbd|)}{\eta^{\mathrm{bd}}}
    +\eta^{\mathrm{bd}}\sum_t(z^{\mathrm{bd}}_{t,f,\ell})^2.
  \]
  The learning rate is chosen as
  \begin{equation}
    \eta^{\mathrm{bd}}=\min\cbr{\frac14,\sqrt{\frac{\log(3e|\calLbd|\,|\calF|)}{4T}}}.
    \label{eq:bounded-rate}
  \end{equation}
  Together with $|z_{t,f,\ell}^{\mathrm{bd}}|\le2$, the preceding comparison yields
  \begin{equation}
    \sum_t z_{t,f,\ell}^{\mathrm{bd}}
    \le\min\cbr{2T,\frac{1+\log(3|\calLbd|\,|\calF|)}{\eta^{\mathrm{bd}}}+4T\eta^{\mathrm{bd}}}
    \le C\sqrt{T\log(e|\calLbd|\,|\calF|)}.
    \label{eq:bounded-correlation}
  \end{equation}
  The last inequality uses \pref{eq:bounded-rate}, with the trivial bound $2T$ when the logarithm is comparable to $T$ or larger.  By \pref{eq:regret_identity_dropping_d}, this also bounds the cumulative loss comparison with every $f\in\calF$, proving \pref{eq:unified-bounded}.  All three guarantees hold simultaneously for the common sequence of forecast distributions.
\end{proof}

\subsection{Consequences for the five stated loss families}
\label{app:consequences}

\unifiedCor*

\begin{proof}[{Proof of \protect\pref{cor:unified}}]
  A single choice of the finite input families suffices.  The finite mixable family contains squared loss and log loss, whose optimal mixability constants are $2$ and $1$, respectively.  The bounded family contains the $0$-$1$ loss in the statement, all the auxiliary losses in \pref{eq:glm-auxiliary-losses}, and their action-reversed versions
  \[
    \ell^-_{k,j}(q,0)=\theta_j\Ind{q<k/T},
    \qquad
    \ell^-_{k,j}(q,1)=(1-\theta_j)\Ind{q\ge k/T}.
  \]

  There are at most $4T(T+1)+1$ bounded losses. We restrict the EVI solver in \pref{alg:unified} to forecast distributions supported on $\Gamma_T=\{i\delta:1\le i\le2T+1\}$, where $\delta=1/[2(T+1)]$. This grid contains both endpoints of $I_\delta$ and has adjacent gaps equal to $\delta$, so \pref{lem:interior-evi} supplies the EVI condition with the same tolerance.
  The first, second, and fifth claims now follow from \pref{thm:unified}: the two mixable losses have $O(\log(e|\calF|))$ regret, the universal mixable bound is retained, and the $0$-$1$ loss has $O(\sqrt{T\log(e|\calF|T)})$ regret.

  \paragraph{The GLM consequence.}
  Fix a continuous convex function $A$ and $f\in\calF$. By \pref{prop:glm-threshold-approximation}, there are coefficients $c_{k,j}\ge0$ with $\sum_{k,j}c_{k,j}=1$ such that \pref{eq:glm-pointwise} holds for every forecast in $\Gamma_T$. Substituting $q=f(x_t)$, taking expectations over $P\sim D_t$, and summing over the rounds gives
  \[
    \begin{aligned}
      &\sum_{t=1}^T\E_{P\sim D_t}
      [\ell_A(\pi_A(P),y_t)-\ell_A(f(x_t),y_t)]\\
      &\quad\le
      \sum_{k=1}^T\sum_{j=1}^{2T+2}
      c_{k,j}\Reg_{\ell_{k,j},\calF}(T)
      +\frac{T\delta}{2}+1\\
      &\quad\le
      C\sqrt{T\log(e|\calF||\calLbd|)}
      +\frac{T\delta}{2}+1.
    \end{aligned}
  \]
  The coefficients are the same in every round, and each cumulative auxiliary loss difference against $f$ is at most its regret against $\calF$. The last inequality therefore follows from \pref{thm:unified} and $\sum_{k,j}c_{k,j}=1$. Since $T\delta/2+1<2$ and $|\calLbd|\le4T(T+1)+1$, taking the best $f\in\calF$ proves the fourth claim, with regret $\order(\sqrt{T\log(e|\calF|T)})$ for every GLM loss.

  It remains to establish the third claim for arbitrary $L$-Lipschitz losses. 

  \paragraph{The Lipschitz consequence.}
  We use the threshold-basis construction of \cite[Lemma~5.4]{okoroafor2025near} for the outcome difference $g(q)=\ell(q,0)-\ell(q,1)=-\Delta_\ell(q)$, and express the resulting approximation using the auxiliary losses above. Both partial losses are $L$-Lipschitz, so $g$ is absolutely continuous and $V=\int_0^1|g'(s)|\dmeas s\le2L$. If $V=0$, all reports have the same outcome difference, and a Bayes action minimizes both partial losses, giving nonpositive regret. For $V>0$, we write
  \[
    \underline g=g(0)-\int_0^1[-g'(s)]_+\dmeas s,
    \qquad
    \psi(p)=\frac{g(\pi_\ell(p))-\underline g}{V}.
  \]
  These definitions give $\underline g\le g(q)\le\underline g+V$ for all $q\in[0,1]$, so $0\le\psi\le1$. Bayes optimality implies that $g(\pi_\ell(p))$ is nondecreasing in $p$. Indeed, adding the optimality inequalities at $p<p'$ gives
  \[
    (p'-p)\sbr{g(\pi_\ell(p'))-g(\pi_\ell(p))}\ge0.
  \]
  Thus $\psi$ is also nondecreasing.
  Let $J=2T+1$, $p_i=i\delta$, and let $a_i=\psi(p_i)$ for $1\le i\le J$, with $a_0=0$ and $a_{J+1}=1$.  The nonnegative numbers $\omega_j=a_j-a_{j-1}$ sum to one and satisfy $\psi(p)=\sum_{j=1}^{J+1}\omega_j\Ind{p>\theta_j}$ for $p\in\Gamma_T$.
  For $1\le k\le T$, let
  \[
    d_k^+=\int_{(k-1)/T}^{k/T}[g'(s)]_+\dmeas s,
    \qquad
    d_k^-=\int_{(k-1)/T}^{k/T}[-g'(s)]_+\dmeas s.
  \]
  Their total mass is $V$. In these two indicator orientations, the threshold approximation on the mesh $1/T$ is
  \[
    \left|g(q)-\underline g-\sum_{k=1}^T
    \sbr{d_k^+\Ind{q\ge k/T}+d_k^-\Ind{q<k/T}}\right|
    \le\frac{2L}{T}
    \qquad(q\in[0,1]).
  \]
  With $\ell^+_{k,j}=\ell_{k,j}$ for the original auxiliary loss, the two orientations have coefficients
  \[
    b_{\ell^+_{k,j}}(p,q)=\Ind{q\ge k/T}-\Ind{p>\theta_j},
    \qquad
    b_{\ell^-_{k,j}}(p,q)=\Ind{q<k/T}-\Ind{p>\theta_j}
  \]
  on the forecast grid.  Since $b_\ell(p,q)=g(q)-g(\pi_\ell(p))$, we obtain
  \[
    \left|b_\ell(p,q)-\sum_{k=1}^T\sum_{j=1}^{J+1}
    \omega_j\sbr{d_k^+b_{\ell^+_{k,j}}(p,q)+d_k^-b_{\ell^-_{k,j}}(p,q)}\right|
    \le\frac{2L}{T}.
  \]
  All coefficients are nonnegative, independent of the round, and have total mass $V\le2L$.  The bounded-component estimate \pref{eq:bounded-correlation} controls each auxiliary correlation by $C\sqrt{T\log(e|\calF||\calLbd|)}$.  After multiplication by $y_t-P$, summation, and forecast expectation, the approximation costs at most $2L$.  The loss comparison in \pref{eq:regret_identity_dropping_d} therefore yields
  \[
    \Reg_{\ell,\calF}(T)\le2CL\sqrt{T\log(e|\calF||\calLbd|)}+2L
    =O\rbr{L\sqrt{T\log(e|\calF|T)}}.
  \]
  This proves the remaining claim.  Since the finite families and the forecasts do not depend on the target $L$, $\ell$, or GLM potential, the five guarantees hold simultaneously.
\end{proof}

\section{Auxiliary Analysis}
\label{app:auxiliary}

\label{app:factors}

\begin{proof}[{Proof of \protect\pref{lem:mixable-comparison}}]
  The parameters $0<\eta\le1/2$, $\beta>0$, $\ell\in\MixLosses{\beta}$, and $f\in\calF$ are arbitrary and fixed. By the definition of $z_{t,f,v}$, the weight update in \pref{alg:mixable} and Jensen's inequality give
  \[
    \begin{aligned}
      \frac{W_{t+1,f}(v)}{W_{t,f}(v)}
      &=\exp\rbr{\E_{P\sim D_t}\sbr{
        \eta b_v(P,f(x_t))(y_t-P)
        -\eta^2P(1-P)b_v(P,f(x_t))^2}}\\
      &\le\E_{P\sim D_t}\sbr{
        F^\eta_{P,y_t}(b_v(P,f(x_t)))}.
    \end{aligned}
  \]
  By \pref{lem:bernoulli-factor} and \pref{eq:corrected-threshold-test},
  \[
    0<F^\eta_{p,y}(b_v(p,q))
    \le1+\widetilde b^\eta_v(p,q)(y-p).
  \]
  Averaging with the normalized weights in \pref{eq:mixable-test} shows that $1+G_t(p)(y-p)\ge0$. Since $\delta/(1-\delta)=1/(2T+1)<\epsilon_t=1/T$, \pref{lem:interior-evi} supplies the required forecast distribution. Applying the same factor bound to the update inequality above, multiplying by $W_{t,f}(v)$, and averaging over $\pi_\delta$ therefore gives, by the definition of $G_t$ and the EVI condition,
  \[
    \Phi_{t+1}
    \le\Phi_t\rbr{1+\E_{P\sim D_t}[G_t(P)(y_t-P)]}
    \le(1+\epsilon_t)\Phi_t.
  \]
  Since $\Phi_1=1$ and $\sum_{t=1}^T\epsilon_t=1$, iteration gives $\Phi_{T+1}\le\prod_{t=1}^T(1+\epsilon_t)\le e$.

  The curvature bound gives $m=\int_{I_\delta}w_\ell(v)\dmeas v\le M_\delta/\beta$. If $m=0$, both threshold integrals in the claimed inequality vanish. For $m>0$, we use $Q=\delta_f\otimes\nu_\ell$, where $\nu_\ell(\dmeas v)=w_\ell(v)\dmeas v/m$, so that averaging over $Q$ gives the desired weighting by $w_\ell$. The prior comparison in \pref{eq:prior-comparison}, applied with $\pi_\delta$, yields
  \[
    \int_{I_\delta}w_\ell(v)\log W_{T+1,f}(v)\dmeas v
    =m\E_Q[\log W_{T+1}]
    \le m\rbr{\KL(Q\|\pi_\delta)+1}.
  \]
  Because the weights start at one, their logarithms are the sums of the update exponents. Substitution into this inequality gives
  \[
    \begin{aligned}
      &\eta\sum_{t=1}^T\int_{I_\delta}w_\ell(v)z_{t,f,v}\dmeas v-\eta^2\sum_{t=1}^T\E_{P\sim D_t}
        \int_{I_\delta}w_\ell(v)P(1-P)b_v(P,f(x_t))^2\dmeas v
      \le m\rbr{\KL(Q\|\pi_\delta)+1}.
    \end{aligned}
  \]
  After division by $\eta$, the second term gives the variance correction in the statement. It remains to bound the right-hand side.

  The prior $\pi_\delta=\operatorname{Unif}(\calF)\otimes\mu_\delta$ assigns mass $1/|\calF|$ to the fixed predictor $f$, and $\mu_\delta(\dmeas v)=\dmeas v/[M_\delta v(1-v)]$. Thus
  \[
    \begin{aligned}
      \KL(Q\|\pi_\delta)
      &=\log|\calF|+\KL(\nu_\ell\|\mu_\delta)\\
      &=\log|\calF|+\int_{I_\delta}
        \log\frac{M_\delta v(1-v)w_\ell(v)}m\,\nu_\ell(\dmeas v)\\
      &\le\log|\calF|+\log\frac{M_\delta}{\beta m},
    \end{aligned}
  \]
  where the last inequality uses $w_\ell(v)\le1/[\beta v(1-v)]$. With $s=\beta m/M_\delta\in(0,1]$, we obtain
  \[
    \begin{aligned}
      m\rbr{\KL(Q\|\pi_\delta)+1}
      &\le m\log(e|\calF|)+m\log\frac{M_\delta}{\beta m}\\
      &=\frac{M_\delta}{\beta}
        \sbr{s\log(e|\calF|)+s\log(1/s)}\\
      &\le\frac{M_\delta}{\beta}
        \sbr{\log(e|\calF|)+1/e},
    \end{aligned}
  \]
  using $s\le1$ and $s\log(1/s)\le1/e$. Substitution and division by $\eta$ give the comparison bound. Its complexity term satisfies
  \[
    \frac{M_\delta[\log(e|\calF|)+1/e]}{\beta\eta}
    =\order\rbr{\frac{\log(T)\log(|\calF|)}{\beta\eta}},
  \]
  since $M_\delta=2\log(2T+1)$. For $T\ge2$ and $|\calF|\ge2$, this is the $\log T\log|\calF|/(\beta\eta)$ dependence stated in the lemma.
\end{proof}

\begin{proof}[{Proof of \protect\pref{lem:mixable-cancellation}}]
  The parameters $\beta,\ell,\eta,p,q$ are arbitrary and fixed as in the statement.  With $h=\eta p(1-p)$, the positive part is confined to $|v-p|<h$, where $v\ge(1-\eta)p$ and $1-v\ge(1-\eta)(1-p)$.  Thus $v(1-v)\ge p(1-p)/4$, and the curvature bound gives
  \[
    \begin{aligned}
      \int_{I_\delta}w_\ell(v)\sbr{\eta p(1-p)b_v(p,q)^2-d_v(p,q)}\dmeas v
      &\le\frac4{\beta p(1-p)}\int_{-h}^h[h-|s|]_+\dmeas s\\
      &=\frac{4\eta^2p(1-p)}\beta\le\frac{\eta^2}{\beta}.\qedhere
    \end{aligned}
  \]
\end{proof}

\begin{proof}[{Proof of \protect\pref{lem:target-clipping}}]
  The parameters $\beta,\ell,q,y$ are arbitrary and fixed as in the statement, and $\bar q=\ClipOp(q,I_\delta)$. If $q\in I_\delta$, clipping leaves the report unchanged, so its cost is zero. For reports outside $I_\delta$, the representation in \pref{lem:automatic-curvature} shows that $\ell(\cdot,0)$ is nondecreasing and $\ell(\cdot,1)$ is nonincreasing.

  When $q<\delta$, we have $\bar q=\delta$, and monotonicity gives $\ell(\delta,1)\le\ell(q,1)$. To bound the increase for outcome $0$, applying \pref{eq:mixability-factor} at report $p=\delta$ gives
  \[
    \begin{aligned}
      &(1-\delta)\exp\rbr{\beta[\ell(\delta,0)-\ell(q,0)]}+\delta\exp\rbr{\beta[\ell(\delta,1)-\ell(q,1)]}\le1.
    \end{aligned}
  \]
  Both terms are nonnegative, so the first is at most one. Taking logarithms yields
  \[
    \ell(\delta,0)-\ell(q,0)
    \le-\frac{\log(1-\delta)}\beta
    \le\frac{\delta}{\beta(1-\delta)}
    \le\frac{2\delta}\beta,
  \]
  where the last inequality uses $\delta<1/2$.

  When $q>1-\delta$, we have $\bar q=1-\delta$ and $\ell(1-\delta,0)\le\ell(q,0)$. Applying \pref{eq:mixability-factor} at $p=1-\delta$ now bounds its outcome-$1$ term by one, giving $(1-\delta)\exp\rbr{\beta[\ell(1-\delta,1)-\ell(q,1)]}\le1$. The same logarithmic bound therefore controls the increase for outcome $1$. If $\ell(q,y)=+\infty$, the corresponding loss difference is $-\infty$, so the claimed bound also holds.

  For the second claim, $p\in I_\delta$ is arbitrary and fixed. For every $v\in(\delta,1-\delta)$, clipping does not change whether $q$ lies below $v$. Hence $b_v(p,q)=b_v(p,\bar q)$, and $d_v(p,q)=d_v(p,\bar q)$ follows from $d_v(p,q)=|v-p|b_v(p,q)^2$. The possible differences at the two endpoint thresholds do not affect the integrals. Since $p,\bar q\in I_\delta$, only thresholds in $I_\delta$ contribute to their loss difference in \pref{eq:local-v-representation}. Combining this representation with the clipping bound gives
  \[
    \begin{aligned}
      \ell(p,y)-\ell(q,y)
      &=\ell(p,y)-\ell(\bar q,y)+\ell(\bar q,y)-\ell(q,y)\\
      &\le\int_{I_\delta}w_\ell(v)
      \sbr{b_v(p,q)(y-p)-d_v(p,q)}\dmeas v
      +\frac{2\delta}\beta,
    \end{aligned}
  \]
  which proves \pref{eq:truncated-comparison}.

  Finally, the corrected test in \pref{eq:corrected-threshold-test} depends on $q$ only through $b_v(p,q)$, and the weight update in \pref{alg:mixable} uses only $b_v(p,q)$ and its square. Thus clipping the benchmark reports leaves the weight densities unchanged almost everywhere and the EVI tests unchanged.
\end{proof}


\section{Preliminary Facts}
\label{app:prelim}

\begin{proof}[{Proof of \protect\pref{lem:automatic-curvature}}]
  \textbf{The representation in \pref{eq:v-shaped-rep}.} Let $\ell$ be a Lipschitz or mixable proper loss, and recall that $\Delta_\ell(p)=\ell(p,1)-\ell(p,0)$. We first show that $\Delta_\ell$ is Lipschitz on every closed subinterval of $(0,1)$, which will allow us to construct the density and recover the partial losses by integration. Properness at $p$ and $q$ gives, for all $0<p<q<1$,
  \[
    p\sbr{\Delta_\ell(p)-\Delta_\ell(q)}
    \le\ell(q,0)-\ell(p,0)
    \le q\sbr{\Delta_\ell(p)-\Delta_\ell(q)}.
  \]
  Hence $\Delta_\ell$ is nonincreasing. If $\ell$ is $L$-Lipschitz, its two partial losses are $L$-Lipschitz, so their difference $\Delta_\ell$ is $2L$-Lipschitz.

  If $\ell\in\MixLosses{\beta}$, concavity of the logarithm applied to \pref{eq:mixability-factor}, with comparator $r$ and weights $1-r$ and $r$, gives, for all $r,p\in(0,1)$,
  \[
    \beta d_\ell(r,p)
    \le\KL\rbr{\Ber(r)\Vert\Ber(p)}.
  \]
  For all $0<p<q<1$, the definition of $d_\ell$ gives
  \[
    d_\ell(q,p)+d_\ell(p,q)
    =(q-p)\sbr{\Delta_\ell(p)-\Delta_\ell(q)},
  \]
  since the terms $\ell(p,0)$ and $\ell(q,0)$ cancel. The definition of Bernoulli KL divergence also gives
  \[
    \begin{aligned}
      \KL(\Ber(q)\Vert\Ber(p))+\KL(\Ber(p)\Vert\Ber(q))
      &=(q-p)\log\frac{q(1-p)}{p(1-q)}\\
      &=(q-p)\int_p^q\frac{\dmeas s}{s(1-s)}.
    \end{aligned}
  \]
  Adding the preceding KL bounds in the two directions and dividing by $\beta(q-p)>0$ therefore yields
  \[
    0\le\Delta_\ell(p)-\Delta_\ell(q)
    \le\frac1\beta\int_p^q\frac{\dmeas s}{s(1-s)}.
  \]
  Since $1/[s(1-s)]$ is bounded on every closed subinterval of $(0,1)$, this inequality makes $\Delta_\ell$ Lipschitz on each such interval.

  Thus, in either case, $\Delta_\ell$ is differentiable almost everywhere. We now define $w_\ell(p):=-\Delta_\ell'(p)$ wherever the derivative exists, and $w_\ell(p)=0$ elsewhere. This gives a nonnegative measurable function because $\Delta_\ell$ is nonincreasing. On every closed subinterval of $(0,1)$, the Lipschitz bound for $\Delta_\ell$ and the properness inequalities above also show that $\ell(\cdot,0)$ is Lipschitz. The identity $\ell(\cdot,1)=\ell(\cdot,0)+\Delta_\ell$ gives the same property for $\ell(\cdot,1)$. Dividing the properness inequalities by $q-p$ and taking the limit gives the first identity below, while differentiating the relation between the partial losses gives the second:
  \[
    \partial_p\ell(p,0)=p w_\ell(p),
    \qquad
    \partial_p\ell(p,1)=-(1-p)w_\ell(p)
    \quad\text{almost everywhere}.
  \]
  On each such interval, Lipschitz continuity ensures that integrating these derivatives recovers the differences of the partial losses. The derivative signs and the endpoint convention in \pref{sec:prelim} imply that $\ell(0,0)$ and $\ell(1,1)$ are finite: each is at most an interior loss value, and the endpoint convention excludes $-\infty$. Passing to these endpoint limits gives
  \[
    \ell(p,0)=\ell(0,0)+\int_0^p v w_\ell(v)\dmeas v,
    \qquad
    \ell(p,1)=\ell(1,1)+\int_p^1(1-v)w_\ell(v)\dmeas v.
  \]
  The equalities extend to $p=0,1$ by the same endpoint convention, allowing an integral to equal $+\infty$. By the definition of $\lambda_v$, the integral for outcome $0$ is $\int_0^1w_\ell(v)\lambda_v(p,0)\dmeas v$, and the integral for outcome $1$ is $\int_0^1w_\ell(v)\lambda_v(p,1)\dmeas v$. Thus, with $C_\ell(y)=(1-y)\ell(0,0)+y\ell(1,1)$,
  \[
    \ell(p,y)=C_\ell(y)+\int_0^1w_\ell(v)\lambda_v(p,y)\dmeas v.
  \]
  This proves \pref{eq:v-shaped-rep}.

  \textbf{The Lipschitz equivalence.} For the remaining claims, let $\ell$ be any proper loss
admitting \pref{eq:v-shaped-rep}.
Since its partial losses are finite on $(0,1)$,
the density $w_\ell$ is integrable on every closed
subinterval of $(0,1)$.
Differentiating the representation gives
$\partial_p\ell(p,0)=p w_\ell(p)$ and
$\partial_p\ell(p,1)=-(1-p)w_\ell(p)$ almost everywhere.
Thus $w_\ell=-\Delta_\ell'$ almost everywhere.

  If $\ell\in\LipLosses{L}$, both partial-loss derivatives have absolute value at most $L$. Thus $p w_\ell(p)\le L$ and $(1-p)w_\ell(p)\le L$, which give $w_\ell(p)\le L/\max\{p,1-p\}$ almost everywhere.

  Conversely, if $w_\ell(v)\le L/\max\{v,1-v\}$ almost everywhere, then $v w_\ell(v)\le L$ and $(1-v)w_\ell(v)\le L$. For all $0\le p<q\le1$, the representation gives
  \[
    \begin{aligned}
      0\le\ell(q,0)-\ell(p,0)
      &=\int_p^q v w_\ell(v)\dmeas v
      \le L(q-p),\\
      0\le\ell(p,1)-\ell(q,1)
      &=\int_p^q(1-v)w_\ell(v)\dmeas v
      \le L(q-p).
    \end{aligned}
  \]
  Thus both partial losses are $L$-Lipschitz.

  \textbf{The mixability equivalence.} Fix $\beta>0$. If $\ell\in\MixLosses{\beta}$, the bound on $\Delta_\ell(p)-\Delta_\ell(q)$ established in the first part gives
  \[
    0\le\frac{\Delta_\ell(p)-\Delta_\ell(q)}{q-p}
    \le\frac1{\beta(q-p)}\int_p^q\frac{\dmeas s}{s(1-s)}.
  \]
  Taking $q\downarrow p$ (i.e., $q$ approaches $p$ from above) and using $w_\ell=-\Delta_\ell'$ gives $w_\ell(p)\le1/[\beta p(1-p)]$ almost everywhere.

  Conversely, suppose $w_\ell(s)\le1/[\beta s(1-s)]$ almost everywhere. For all $r,p\in(0,1)$, the definition of $d_\ell$ gives $d_\ell(r,p)=(1-r)[\ell(p,0)-\ell(r,0)]+r[\ell(p,1)-\ell(r,1)]$. The derivative identities above imply $(1-r)\partial_s\ell(s,0)+r\partial_s\ell(s,1)=(s-r)w_\ell(s)$ almost everywhere. Integrating from $r$ to $p$ and using the bound on $w_\ell$ therefore yields
  \[
    \beta d_\ell(r,p)
    =\beta\int_r^p(s-r)w_\ell(s)\dmeas s
    \le\int_r^p\frac{s-r}{s(1-s)}\dmeas s
    =\KL\rbr{\Ber(r)\Vert\Ber(p)}.
  \]
  Consequently, for all $p,q\in(0,1)$,
  \[
    \log\E_{Y\sim\Ber(p)}\exp\rbr{\beta[\ell(p,Y)-\ell(q,Y)]}
    =\sup_{r\in(0,1)}\cbr{\beta[d_\ell(r,p)-d_\ell(r,q)]-\KL(\Ber(r)\Vert\Ber(p))}\le0,
  \]
  where the equality follows by maximizing over $r$, and the inequality uses the preceding bound and $d_\ell(r,q)\ge0$.

  For any $0<\varepsilon<1/2$ and any distribution $D$ supported on $[\varepsilon,1-\varepsilon]$, continuity of $q\mapsto\exp\rbr{-\beta\ell(q,0)}$ gives a report $p\in[\varepsilon,1-\varepsilon]$ satisfying
  \[
    \exp\rbr{-\beta\ell(p,0)}
    =\E_{Q\sim D}\exp\rbr{-\beta\ell(Q,0)}.
  \]
  Exponentiating the logarithmic bound above and averaging over $q\sim D$ then gives
  \[
    \exp\rbr{-\beta\ell(p,1)}
    \ge\E_{Q\sim D}\exp\rbr{-\beta\ell(Q,1)},
  \]
  since the outcome-$0$ term in the averaged inequality equals $1-p$. For an arbitrary distribution $D$ on $[0,1]$, the same argument applies after clipping its reports to $[\varepsilon,1-\varepsilon]$. As $\varepsilon\downarrow0$, compactness of $[0,1]$ gives a convergent subsequence of the resulting reports. The integral representation makes $q\mapsto\exp\rbr{-\beta\ell(q,y)}$ bounded and continuous on $[0,1]$ for both outcomes, with $\exp\rbr{-\infty}=0$. Bounded convergence for the clipped reports therefore implies that the limiting report satisfies both inequalities for the original $D$. This is $\beta$-mixability.

  \textbf{The consequence for Lipschitz losses.} Finally, if $\ell\in\LipLosses{L}$, the Lipschitz curvature bound gives
  \[
    v(1-v)w_\ell(v)
    \le L\min\{v,1-v\}\le L/2
    \quad\text{almost everywhere}.
  \]
  The sufficient condition for mixability with $\beta=2/L$ therefore gives $\ell\in\MixLosses{2/L}$.
\end{proof}


\begin{proof}[{Proof of \protect\pref{lem:interior-evi}}]
  The parameters $\delta,G,\epsilon$ are arbitrary and fixed as in the statement. The assumption $1+G(p)(y-p)\ge0$ at $y=0$ and $y=1$ gives $1-pG(p)\ge0$ and $1+(1-p)G(p)\ge0$, respectively. Hence, for every $p\in I_\delta\cap(0,1)$,
  \[
    -\frac1{1-p}\le G(p)\le\frac1p.
  \]

  We use a finite grid $\Gamma\subseteq I_\delta$, meaning a finite set of possible forecast values, containing both endpoints of $I_\delta$. The gaps between consecutive values will be specified below. We first show that, for every $r\in[0,1]$, some $p\in\Gamma$ satisfies $G(p)(r-p)\le\epsilon$. Here $r$ is the probability of outcome $1$, and
  \[
    \E_{Y\sim\Ber(r)}[G(p)(Y-p)]=G(p)(r-p).
  \]
  The minimax theorem will then give a single distribution on $\Gamma$ that satisfies the desired bound for both outcomes.

  The value $r\in[0,1]$ is arbitrary and fixed. If $r\in\Gamma$, then $r$ itself is an allowed forecast, and $p=r$ gives $G(p)(r-p)=0$. Two grid points are consecutive if no point of $\Gamma$ lies strictly between them. When $r$ lies between consecutive grid points $p_-,p_+\in\Gamma$ satisfying $0<p_-<r<p_+<1$, the bounds on $G$ give, with $d=p_+-p_-$,
  \[
    \begin{aligned}
      \min\cbr{G(p_-)(r-p_-),G(p_+)(r-p_+)}
      &\le\min\cbr{\frac{r-p_-}{p_-},\frac{p_+-r}{1-p_+}}\\
      &\le\frac{(r-p_-)+(p_+-r)}{p_-+(1-p_+)}
      =\frac{d}{1-d}.
    \end{aligned}
  \]
  The first inequality uses the upper bound on $G(p_-)$ and the lower bound on $G(p_+)$, since $r-p_->0$ and $r-p_+<0$. For the second inequality, the fraction is a weighted average of the two preceding fractions, with positive weights $p_-$ and $1-p_+$, and is therefore at least their minimum. Thus one of the two forecasts gives a value at most $d/(1-d)$.

  When $\delta>0$, $\Gamma$ is any grid allowed by the statement. All its points lie in $(0,1)$, and every consecutive gap satisfies $d\le\delta$. The preceding argument therefore gives a forecast with value at most $\delta/(1-\delta)<\epsilon$ for every $r\in I_\delta$. If $r$ is outside $I_\delta$, the nearest endpoint gives the same bound. More explicitly,
  \[
    \begin{aligned}
      G(\delta)(r-\delta)
      &\le\frac{\delta-r}{1-\delta}
      \le\frac\delta{1-\delta}
      &&\text{if }r<\delta,\\
      G(1-\delta)(r-1+\delta)
      &\le\frac{r-1+\delta}{1-\delta}
      \le\frac\delta{1-\delta}
      &&\text{if }r>1-\delta.
    \end{aligned}
  \]

  When $\delta=0$, we choose a finite grid $\Gamma$ containing $0$ and $1$ whose consecutive gaps are at most $\epsilon/(1+\epsilon+|G(0)|+|G(1)|)$. Such a grid exists because $G(0)$ and $G(1)$ are finite. This gap bound is at most $\epsilon/(1+\epsilon)$, so $d/(1-d)\le\epsilon$ whenever $r$ lies between consecutive points strictly inside $(0,1)$. If $r$ lies between $0$ and the next grid point, the forecast $p=0$ gives
  \[
    G(0)r\le |G(0)|r
    \le\frac{\epsilon|G(0)|}{1+\epsilon+|G(0)|+|G(1)|}
    \le\epsilon.
  \]
  Similarly, if $r$ lies between the last grid point below $1$ and $1$, the forecast $p=1$ gives
  \[
    G(1)(r-1)\le |G(1)|(1-r)
    \le\frac{\epsilon|G(1)|}{1+\epsilon+|G(0)|+|G(1)|}
    \le\epsilon.
  \]
  Together with the case $r\in\Gamma$, this covers every $r\in[0,1]$.

  In both cases, we have shown that $\min_{p\in\Gamma}G(p)(r-p)\le\epsilon$ for every $r\in[0,1]$. Applying the finite minimax theorem to the forecast choices $p\in\Gamma$ and outcome choices $y\in\{0,1\}$ gives
  \[
    \begin{aligned}
      &\min_{D\text{ on }\Gamma}\max_{y\in\{0,1\}}
      \E_{P\sim D}[G(P)(y-P)] =\max_{r\in[0,1]}\min_{p\in\Gamma}G(p)(r-p)
      \le\epsilon,
    \end{aligned}
  \]
  where the minimum is over probability distributions $D$ on $\Gamma$. Thus the same distribution $D$ satisfies the bound for both $y=0$ and $y=1$.

   It remains to limit $D$ to at most three forecast values.
  The pair of expectations for $y=0$ and $y=1$ is an average,
  with weights given by $D$, of the vectors
  $(-pG(p),(1-p)G(p))$ for $p\in\Gamma$.
  Carath\'eodory's theorem in $\mathbb R^2$ states that this
  average can be expressed using at most three of these vectors
  with nonnegative weights summing to one.
  These weights give a distribution on at most three points
  of $\Gamma$, again denoted by $D$, with the same two
  expectations. Hence the required bound holds for both
  $y=0$ and $y=1$.
\end{proof}

\section{Approximate Bases for GLM Losses}
\label{app:glm}

For an integer $T\ge1$, let $\delta=1/[2(T+1)]$,
$\Gamma_T=\{i\delta:1\le i\le2T+1\}$, and
$\theta_j=(j-\tfrac12)\delta$ for $1\le j\le2T+2$.
For $1\le k\le T$ and $1\le j\le2T+2$, define the auxiliary losses
\begin{equation}
  \ell_{k,j}(q,0)=\theta_j\Ind{q\ge k/T},
  \qquad
  \ell_{k,j}(q,1)=(1-\theta_j)\Ind{q<k/T}.
  \label{eq:glm-auxiliary-losses}
\end{equation}
These $2T(T+1)$ losses take values in $[0,1]$ and admit the
Bayes action $\pi_{\ell_{k,j}}(p)=\Ind{p>\theta_j}$ on $\Gamma_T$.
Both the grid and the auxiliary losses depend only on $T$.

\begin{proposition}[Finite threshold approximation for GLM loss comparisons]
  \label{prop:glm-threshold-approximation}
  For every continuous convex function $A:[0,1]\to\mathbb R$,
  let $\ell_A(q,y)=A(q)-yq$ and fix a Bayes post-processing $\pi_A$.
  There exist coefficients $c_{k,j}\ge0$ with
  $\sum_{k=1}^T\sum_{j=1}^{2T+2}c_{k,j}=1$ such that,
  for every $p\in\Gamma_T$, $q\in[0,1]$, and $y\in\{0,1\}$,
  \begin{equation}
    \begin{aligned}
      \ell_A(\pi_A(p),y)-\ell_A(q,y)\le\sum_{k=1}^T\sum_{j=1}^{2T+2}c_{k,j}
      \sbr{\ell_{k,j}(\pi_{\ell_{k,j}}(p),y)-\ell_{k,j}(q,y)}
      +\frac\delta2+\frac1T.
    \end{aligned}
    \label{eq:glm-pointwise}
  \end{equation}
\end{proposition}

\begin{proof}[{Proof of \protect\pref{prop:glm-threshold-approximation}}]
  Fix a continuous convex function $A$ and its Bayes post-processing $\pi_A$.
  Write $J=2T+1$, $p_i=i\delta$, and $q_i=\pi_A(p_i)$ for
  $1\le i\le J$, where $\delta=1/[2(T+1)]$.
  We also set $p_0=0$, $p_{J+1}=1$, $q_0=0$, and $q_{J+1}=1$
  to include the endpoints in the partitions below.

  \textbf{Approximating the loss differences.}
  For $i<i'$, Bayes optimality gives
  \[
    \begin{aligned}
      A(q_i)-p_iq_i
      &\le A(q_{i'})-p_iq_{i'},\\
      A(q_{i'})-p_{i'}q_{i'}
      &\le A(q_i)-p_{i'}q_i.
    \end{aligned}
  \]
  Adding these inequalities yields
  $(p_{i'}-p_i)(q_{i'}-q_i)\ge0$.
  Since $p_{i'}>p_i$, we obtain $q_i\le q_{i'}$.
  Thus $q_0\le\cdots\le q_{J+1}$, and the intervals
  $(q_{j-1},q_j]$ partition $(0,1]$.

  Bayes optimality also gives
  $A(q)\ge A(q_i)+p_i(q-q_i)$ for every $q\in[0,1]$.
  We therefore define
  \[
    \bar A(q)
    =\max_{1\le i\le J}\{A(q_i)+p_i(q-q_i)\}.
  \]
  Each line in this maximum lies below $A$, so $\bar A\le A$.
  At $q_i$, the $i$th line equals $A(q_i)$, giving
  $\bar A(q_i)=A(q_i)$.
  Consequently, for any comparator report $q\in[0,1]$
  and outcome $y\in\{0,1\}$,
  \[
    \ell_A(q_i,y)-\ell_A(q,y)
    \le\bar A(q_i)-\bar A(q)-y(q_i-q).
  \]

  We next bound the slopes of $\bar A$ between consecutive $q_i$.
  For $2\le j\le J$, consider a nonempty interval
  $(q_{j-1},q_j)$.
  At $q_{j-1}$, every line in the maximum is at most
  $A(q_{j-1})$.
  The lines indexed by $i\le j-1$ have slopes at most $p_{j-1}$,
  so to the right of $q_{j-1}$ they are no larger than
  the line indexed by $j-1$.
  Similarly, to the left of $q_j$, the lines indexed by $i\ge j$
  are no larger than the line indexed by $j$.
  Hence $\bar A$ is the maximum of these two lines on
  $(q_{j-1},q_j)$, and its derivative lies between
  $p_{j-1}$ and $p_j$ wherever it exists.
  On $(0,q_1)$ the first line attains the maximum,
  and on $(q_J,1)$ the last line attains the maximum.
  Thus, for every $1\le j\le J+1$,
  \[
    p_{j-1}\le\bar A'(s)\le p_j
    \quad\text{for almost every }s\in(q_{j-1},q_j).
  \]
  Since $\theta_j=(p_{j-1}+p_j)/2$ and $p_j-p_{j-1}=\delta$,
  this gives $|\bar A'(s)-\theta_j|\le\delta/2$ on that interval.

  The function $\bar A$ is continuous and piecewise linear,
  so integrating its derivative gives
  \[
    \begin{aligned}
      &\bar A(q_i)-\bar A(q)-y(q_i-q)\\
      &\quad=
      \sum_{j=1}^{J+1}\int_{q_{j-1}}^{q_j}
      (\bar A'(s)-y)
      [\Ind{s\le q_i}-\Ind{s\le q}]\dmeas s\\
      &\quad\le
      \sum_{j=1}^{J+1}(\theta_j-y)
      \int_{q_{j-1}}^{q_j}
      [\Ind{s\le q_i}-\Ind{s\le q}]\dmeas s
      +\frac\delta2.
    \end{aligned}
  \]
  Indeed, the indicator difference is nonzero only between $q_i$
  and $q$, where its absolute value is one.
  Replacing $\bar A'(s)$ by $\theta_j$ therefore introduces
  an error at most $(\delta/2)|q_i-q|\le\delta/2$.

  Let $q^-=\lfloor Tq\rfloor/T$.
  Replacing $\Ind{s\le q}$ by $\Ind{s\le q^-}$ changes the
  integrand only on $(q^-,q]$, whose length is at most $1/T$.
  Since $|\theta_j-y|\le1$, this introduces an additional
  error at most $1/T$.

  \textbf{Expressing the approximation using the auxiliary losses.}
  For $1\le k\le T$ and $1\le j\le J+1$, let
  \[
    c_{k,j}
    =\Leb\bigl(((k-1)/T,k/T]\cap(q_{j-1},q_j]\bigr).
  \]
  Thus $c_{k,j}$ is the length of the intersection of the two
  indicated intervals.
  Both collections of intervals partition $(0,1]$, so
  $c_{k,j}\ge0$ and $\sum_{k,j}c_{k,j}=1$.
  These coefficients depend only on $A$, its fixed
  post-processing $\pi_A$, and the grids.

  On each nonempty intersection, the two indicators in the
  preceding approximation are constant.
  Since $q^-$ is a multiple of $1/T$, we have
  $\Ind{s\le q^-}=\Ind{q\ge k/T}$ there.
  Likewise, the ordering of the $q_i$ gives
  $\Ind{s\le q_i}=\Ind{j\le i}$.
  Because $p_i=i\delta$ and $\theta_j=(j-\tfrac12)\delta$,
  this last indicator equals $\Ind{p_i>\theta_j}$.
  Splitting the integrals over these intersections therefore gives
  \[
    \begin{aligned}
      \sum_{j=1}^{J+1}(\theta_j-y)
      \int_{q_{j-1}}^{q_j}
      [\Ind{s\le q_i}-\Ind{s\le q^-}]\dmeas s=
      \sum_{k=1}^T\sum_{j=1}^{J+1}c_{k,j}(\theta_j-y)
      [\Ind{p_i>\theta_j}-\Ind{q\ge k/T}].
    \end{aligned}
  \]

  To identify these terms as auxiliary loss differences,
  the definition in \pref{eq:glm-auxiliary-losses} gives
  \[
    \ell_{k,j}(q,y)
    =(1-\theta_j)y+(\theta_j-y)\Ind{q\ge k/T}.
  \]
  Its Bayes action on the forecast grid is
  $\pi_{\ell_{k,j}}(p_i)=\Ind{p_i>\theta_j}$.
  This action is either zero or one, and $0<k/T\le1$, so
  \[
    \begin{aligned}
      &\ell_{k,j}(\pi_{\ell_{k,j}}(p_i),y)-\ell_{k,j}(q,y)=
      (\theta_j-y)
      [\Ind{p_i>\theta_j}-\Ind{q\ge k/T}].
    \end{aligned}
  \]
  Combining the preceding identities with the two approximation
  errors proves \pref{eq:glm-pointwise}.
\end{proof}

\bibliographystyle{alpha}
\bibliography{refs}

@string{PMLR = {PMLR}}

@string{PMLR = {Proceedings of Machine Learning Research}}

@article{farina2026blackbox,
  title={An Efficient Black-Box Reduction from Online Learning to Multicalibration, and a New Route to $\Phi$-Regret Minimization},
  author={Farina, Gabriele and Perdomo, Juan Carlos},
  journal={arXiv preprint arXiv:2604.19592},
  year={2026}
}

@inproceedings{zhang2025evi,
  title={Expected Variational Inequalities},
  author={Zhang, Brian Hu and Anagnostides, Ioannis and Tewolde, Emanuel and Berker, Ratip Emin and Farina, Gabriele and Conitzer, Vincent and Sandholm, Tuomas},
  booktitle={Proceedings of the 42nd International Conference on Machine Learning},
  year={2025}
}

@article{okoroafor2025near,
  title={Near-Optimal Algorithms for Omniprediction},
  author={Okoroafor, Princewill and Kleinberg, Robert and Kim, Michael P.},
  journal={Symposium on Foundations of Computer Science},
  year={2025}
}

@article{okoroafor2026swap,
  title={Fast Rates for Swap-Agnostic Learning of Proper Losses},
  author={Okoroafor, Princewill},
  journal={arXiv preprint arXiv:2607.28856},
  year={2026}
}

@inproceedings{gopalan2023swap,
  title={Swap Agnostic Learning, or Characterizing Omniprediction via Multicalibration},
  author={Gopalan, Parikshit and Kim, Michael P. and Reingold, Omer},
  booktitle={Advances in Neural Information Processing Systems},
  volume={36},
  pages={39936--39956},
  year={2023}
}

@inproceedings{hu2026multiclass,
  title={Simultaneous Blackwell Approachability and Applications to Multiclass Omniprediction},
  author={Hu, Lunjia and Tian, Kevin and Yang, Chutong},
  booktitle={Proceedings of the Thirty-Ninth Conference on Learning Theory},
  year={2026}
}

@inproceedings{chernov2009prediction,
  title={Prediction with expert evaluators’ advice},
  author={Chernov, Alexey and Vovk, Vladimir},
  booktitle={International Conference on Algorithmic Learning Theory},
  pages={8--22},
  year={2009},
  organization={Springer}
}

@inproceedings{gopalan2022omnipredictors,
  title={Omnipredictors},
  author={Gopalan, Parikshit and Kalai, Adam Tauman and Reingold, Omer and Sharan, Vatsal and Wieder, Udi},
  booktitle={13th Innovations in Theoretical Computer Science Conference (ITCS 2022)},
  year={2022}
}

@inproceedings{garg2024oracle,
  title={Oracle Efficient Online Multicalibration and Omniprediction},
  author={Garg, Sumegha and Jung, Christopher and Reingold, Omer and Roth, Aaron},
  booktitle={Proceedings of the 2024 Annual ACM-SIAM Symposium on Discrete Algorithms (SODA)},
  pages={2725--2792},
  year={2024},
  publisher={SIAM}
}

@article{chernov2010supermartingales,
  title={Supermartingales in prediction with expert advice},
  author={Chernov, Alexey and Kalnishkan, Yuri and Zhdanov, Fedor and Vovk, Vladimir},
  journal={Theoretical Computer Science},
  volume={411},
  number={29-30},
  pages={2647--2669},
  year={2010},
  publisher={Elsevier}
}

@book{cesa2006prediction,
  title={Prediction, learning, and games},
  author={Cesa-Bianchi, Nicolo and Lugosi, G{\'a}bor and others},
  volume={1},
  number={1.1},
  year={2006},
  publisher={Cambridge university press Cambridge}
}

@article{luo2024optimal,
  title={Optimal multiclass u-calibration error and beyond},
  author={Luo, Haipeng and Senapati, Spandan and Sharan, Vatsal},
  journal={Advances in Neural Information Processing Systems},
  volume={37},
  pages={7521--7551},
  year={2024}
}

@article{luo2025improved,
  title={Improved bounds for swap multicalibration and swap omniprediction},
  author={Luo, Haipeng and Senapati, Spandan and Sharan, Vatsal},
  journal={Advances in Neural Information Processing Systems},
  volume={38},
  pages={20425--20467},
  year={2025}
}

@inproceedings{hebert2018multicalibration,
  title={Multicalibration: Calibration for the (computationally-identifiable) masses},
  author={H{\'e}bert-Johnson, Ursula and Kim, Michael and Reingold, Omer and Rothblum, Guy},
  booktitle={International conference on machine learning},
  pages={1939--1948},
  year={2018},
  organization={PMLR}
}

@article{dewasurendra2026dirichlet,
  title={Dirichlet Follow-the-Leader Closes the Gap in Simultaneous Multiclass U-Calibration},
  author={Dewasurendra, Pahan},
  journal={arXiv preprint arXiv:2608.06656},
  year={2026}
}

@article{kleinberg2023u,
  title={U-calibration: Forecasting for an unknown agent},
  author={Kleinberg, Robert and Leme, Renato Paes and Schneider, Jon and Teng, Yifeng},
  journal={Conference on Learning Theory},
  year={2023}
}

@article{frongillo2026toward,
  title={Toward Simultaneously Optimal Regret in U-Calibration},
  author={Frongillo, Rafael and Luo, Haipeng and Mehta, Nishant A and Schneider, Jon},
  journal={Conference on Learning Theory},
  year={2026}
}

@article{freund1997decision,
  title={A decision-theoretic generalization of on-line learning and an application to boosting},
  author={Freund, Yoav and Schapire, Robert E},
  journal={Journal of computer and system sciences},
  volume={55},
  number={1},
  pages={119--139},
  year={1997},
  publisher={Elsevier}
}

@article{zhang2025expected,
  title={Expected variational inequalities},
  author={Zhang, Brian Hu and Anagnostides, Ioannis and Tewolde, Emanuel and Berker, Ratip Emin and Farina, Gabriele and Conitzer, Vincent and Sandholm, Tuomas},
  journal={arXiv preprint arXiv:2502.18605},
  year={2025}
}

@inproceedings{hoeven2018many,
  title={The many faces of exponential weights in online learning},
  author={Hoeven, Dirk and Erven, Tim and Kot{\l}owski, Wojciech},
  booktitle={Conference On Learning Theory},
  pages={2067--2092},
  year={2018},
  organization={PMLR}
}

@inproceedings{steinhardt2014adaptivity,
  title={Adaptivity and optimism: An improved exponentiated gradient algorithm},
  author={Steinhardt, Jacob and Liang, Percy},
  booktitle={International conference on machine learning},
  pages={1593--1601},
  year={2014},
  organization={PMLR}
}

@article{hazan2010extracting,
  title={Extracting certainty from uncertainty: Regret bounded by variation in costs},
  author={Hazan, Elad and Kale, Satyen},
  journal={Machine learning},
  volume={80},
  number={2},
  pages={165--188},
  year={2010},
  publisher={Springer}
}

@inproceedings{chen2021impossible,
  title={Impossible tuning made possible: A new expert algorithm and its applications},
  author={Chen, Liyu and Luo, Haipeng and Wei, Chen-Yu},
  booktitle={Conference on Learning Theory},
  pages={1216--1259},
  year={2021},
  organization={PMLR}
}

@article{wintenberger2017optimal,
  title={Optimal learning with Bernstein online aggregation},
  author={Wintenberger, Olivier},
  journal={Machine Learning},
  volume={106},
  number={1},
  pages={119--141},
  year={2017},
  publisher={Springer}
}

@article{rakhlin2014statistical,
  title={Statistical learning and sequential prediction},
  author={Rakhlin, Alexander and Sridharan, Karthik},
  journal={Book Draft},
  year={2014}
}

@article{skorski2020handy,
  title={Handy formulas for binomial moments},
  author={Skorski, Maciej},
  journal={arXiv preprint arXiv:2012.06270},
  year={2020}
}

@article{van2015fast,
  title={Fast rates in statistical and online learning},
  author={Van Erven, Tim and Gr{\"u}nwald, Peter and Mehta, Nishant and Reid, Mark D and Williamson, Robert C},
  journal={Journal of Machine Learning Research},
  year={2015}
}

\end{document}